\documentclass[]{camlsys}

\usepackage{graphicx} 
\usepackage{xspace}
\usepackage{amsmath} 
\usepackage[utf8]{inputenc} 
\usepackage[T1]{fontenc}    
\usepackage{hyperref}       
\usepackage{url}            
\usepackage{booktabs}       
\usepackage{amsfonts}       
\usepackage{nicefrac}       
\usepackage{microtype}      
\usepackage{xcolor}         
\usepackage{multirow}
\usepackage{subcaption}
\usepackage{caption}

\title{\texttt{AbbIE}: Autoregressive Block-Based Iterative Encoder for Efficient Sequence Modeling}
\author[1,2]{Preslav Aleksandrov}
\author[1]{Meghdad Kurmanji}
\author[3]{Fernando Garcia Redondo}
\author[1]{David O'Shea}
\author[1]{William Shen}
\author[1]{Alex Iacob}
\author[1]{Lorenzo Sani}
\author[1,2]{Xinchi Qiu}
\author[2]{Nicola Cancedda}
\author[1]{Nicholas D. Lane}

\affiliation[1]{University of Cambridge}
\affiliation[2]{Meta}
\affiliation[3]{IMEC}

\newcommand{\abe}{\texttt{AbbIE}\xspace}
\newcommand{\diff}{\texttt{AbbIE-D}\xspace}
\newcommand{\co}{\texttt{AbbIE-C}\xspace}
\newcommand{\std}{\texttt{Std}\xspace}
\newcommand{\depth}{\texttt{Depth}\xspace}

\begin{document}

\abstract{
We introduce the Autoregressive Block-Based Iterative Encoder (\abe), a novel recursive generalization of the encoder-only Transformer architecture, which achieves better perplexity than a standard Transformer and allows for the dynamic scaling of compute resources at test time. This simple, recursive approach is a complement to scaling large language model (LLM) performance through parameter and token counts. \abe performs its iterations in latent space, but unlike latent reasoning models, does not require a specialized dataset or training protocol. We show that \abe upward generalizes (ability to generalize to arbitrary iteration lengths) at test time by only using \textbf{2} iterations during train time, far outperforming alternative iterative methods. \abe's ability to scale its computational expenditure based on the complexity of the task gives it an up to \textbf{12\%} improvement in zero-shot in-context learning tasks versus other iterative and standard methods and up to \textbf{5\%} improvement in language perplexity. The results from this study open a new avenue to Transformer performance scaling. We perform all of our evaluations on model sizes up to \textbf{350M} parameters.
}

\maketitle


\section{Introduction}
Transformer architectures have become foundational in modern machine learning, enabling state-of-the-art performance across diverse domains \citep{radford2019language, jumper_highly_2021}. The Transformer's performance has been shown to depend on the size of the network (parameter count) \citep{DBLP:journals/corr/abs-2203-15556, DBLP:journals/corr/abs-2001-08361, DBLP:journals/corr/abs-2412-19437}, the amount of data used for training \citep{hägele2024scalinglawscomputeoptimaltraining, sardana2025chinchillaoptimalaccountinginferencelanguage}, and to a degree on the quality of the data the network is trained on \citep{DBLP:journals/corr/abs-2502-02737, DBLP:conf/nips/PenedoKALMRW024}.%

Scaling performance remains challenging due to hardware limitations. In particular, the slower growth of GPU memory capacity compared to computational throughput presents a significant bottleneck to increasing Transformer size for improved performance \citep{wolters2024memoryneedoverviewcomputeinmemory}. Different methods have been developed to alleviate this challenge. Examples include quantizing models into lower precisions \citep{wang20244}, distilling larger models into smaller ones \citep{hsieh2023distilling}, and pruning parts of the network \citep{lu2024alphapruning}. Another popular way to circumvent the memory constraints of Transformers is by using a Mixture of Experts (MoE) architecture \citep{fedus2022switch}, which decreases memory requirements by activating only certain parts of the model during inference \citep{pope2022efficientlyscalingtransformerinference, du2022glamefficientscalinglanguage}. 
\begin{figure}[htbp]
    \centering
    \begin{subfigure}[b]{0.24\textwidth}
        \centering
        \includegraphics[width=\textwidth]{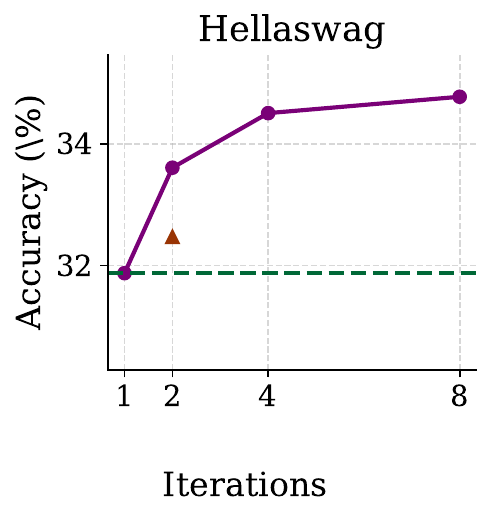}
        \label{fig:subfig1}
    \end{subfigure}
    \hfill
    \begin{subfigure}[b]{0.24\textwidth}
        \centering
        \includegraphics[width=\textwidth]{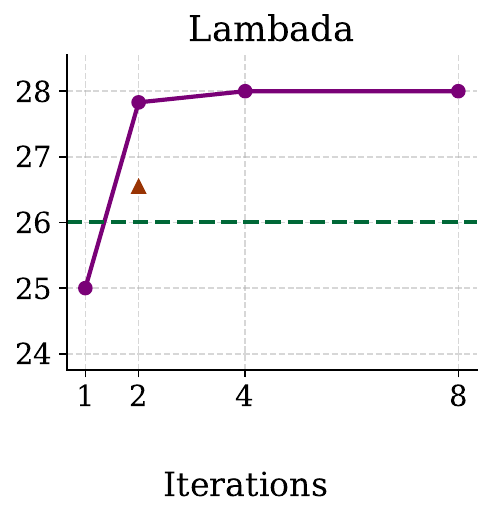}
        \label{fig:subfig1}
    \end{subfigure}
    \hfill
    \begin{subfigure}[b]{0.24\textwidth}
        \centering
        \includegraphics[width=\textwidth]{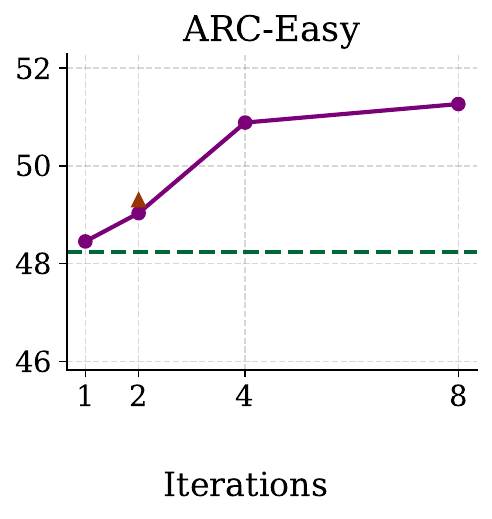}
        \label{fig:subfig1}
    \end{subfigure}
    \hfill
    \begin{subfigure}[b]{0.24\textwidth}
        \centering
        \includegraphics[width=\textwidth]{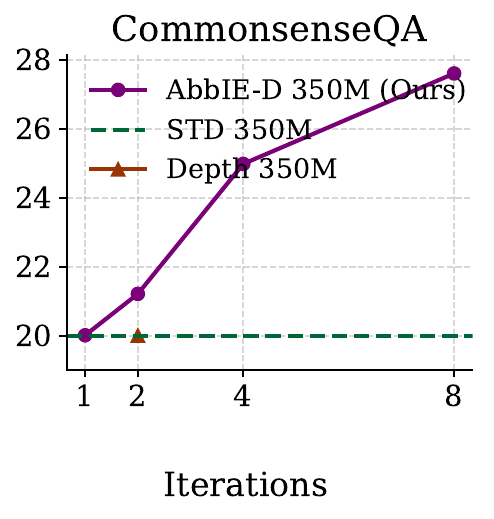}
        \label{fig:subfig1}
    \end{subfigure}
    \vspace{-1em}
    \caption{Performance comparison of \diff (our method) versus baseline models STD and \depth across four benchmarks (Hellaswag, Lambada, ARC-Easy, and CommonsenseQA) for 350M model. Results demonstrate consistent accuracy improvements with increasing training iterations (1, 2, 4, and 8), showing our model significantly outperforms baselines across all benchmarks.}
    \label{fig:ICLScaling}
    \vspace{-1.5em}
\end{figure}

Recently, recurrent formulations of Transformers have appeared as a promising candidate to continue scaling performance without increasing memory \citep{geiping2025scalingtesttimecomputelatent, DBLP:conf/iclr/Yang0NP24}. These formulations have shown that the Transformer can operate several times on its token stream, iteratively, to improve performance \citep{DBLP:conf/iclr/Yang0NP24}. Iteration is employed during both training and inference to allow scaling of resources at test time. This mode of operation is distinctly different from other seemingly recursive language models such as selective state space models (Mamba) \citep{gu2023mamba} or recursive formulations of the attention mechanism (RKWV) \citep{peng2023rwkv}. Recurrent Transformers have also been shown to have superior reasoning capabilities, sometimes outperforming models with an order of magnitude more parameters \citep{DBLP:conf/iclr/Yang0NP24}. However, recurrent Transformers usually rely on a large number of iterations to generalize during training and are often bound to a particular task \citep{bae2024relaxed, DBLP:conf/icml/GiannouRS0LP23, DBLP:conf/iclr/Yang0NP24}. As a result, they are not suitable as drop-in replacements for standard Transformer architectures. The high iteration count also leads to reduced throughput (as the total number of parameters being inferred becomes much larger than the number of unique parameters) and increased training cost. To address these limitations, we propose the \textbf{A}utoregressive \textbf{b}lock-\textbf{b}ased \textbf{I}terative \textbf{E}ncoder (\abe), an iterative formulation of the encoder-only Transformer. \abe leverages the structure of a Transformer's latent space achieve better perplexities and downstream task performance, but unlike latent reasoning models (Coconut) \citep{DBLP:journals/corr/abs-2412-06769} does not require a specialized dataset. Two variants emerge, namely \co and \diff. The difference between the two is in way they handle input injection (Fig.~\ref{fig:abbie_struc}). We study and compare both variants. Unlike previous iterative architectures, \abe is a generalization of the Transformer network. It can be used as a direct replacement: it matches the performance of a standard Transformer after a single iteration, and requires only \textbf{two iterations} during training to enable \textbf{lower perplexity} and \textbf{higher ICL scores} (Fig.~\ref{fig:ICLScaling}) when scaling iterations at test time. This allows \abe to be a viable alternative to standard Transformers even when scaling test time compute is not necessary or possible. 

Through extensive experiments, we show that \abe consistently outperforms standard Transformers for a fixed token budget. When targeting a fixed perplexity at token amounts beyond the compute-optimal point, \abe approaches the FLOP cost of a standard Transformer, whereas previous recursive methods have a significantly higher FLOP cost. This makes \abe a computationally efficient addition to large-scale training runs, as typically modern Transformers train to well above $20\times$ the compute-optimal token count \citep{allal2025smollm2smolgoesbig, DBLP:journals/corr/abs-2412-19437}. We demonstrate that by only training on two iterations, we enable \abe to scale to arbitrary iteration numbers with lower perplexity and improve performance on a range of in-context learning (ICL) tasks during inference. This property allows \abe to match the performance of a standard Transformer, while retaining the flexibility to scale inference compute when beneficial. To our knowledge, \abe is the only recurrent Transformer which possesses these characteristics. They make our method readily adoptable, as it functions equivalently to a standard Transformer, unless test-time scaling is explicitly desired. In summary, the contributions of this paper are:
\begin{itemize}
    \item A new recurrent method for auto-regressive Transformers, termed \abe, that can scale performance during test time.
    \item A training methodology for recurrent Transformers networks, which requires only two iteration at train time.
    \item Extensive experiments on language modeling tasks that show \abe outperforms standard Transformers for the same token budget.
    \item Extensive experiments on zero-shot In-Context Learning (ICL) tasks that show \abe can improved the performance past the number of training iterations.
\end{itemize}

\vspace{-1em}
\section{Autoregressive block-based Iterative Encoder: \abe}
\vspace{-0.8em}
In this section, we present the architecture and operational mechanism of \abe.
\subsection{Design}
\begin{figure}
    \centering
    \includegraphics[width=\linewidth]{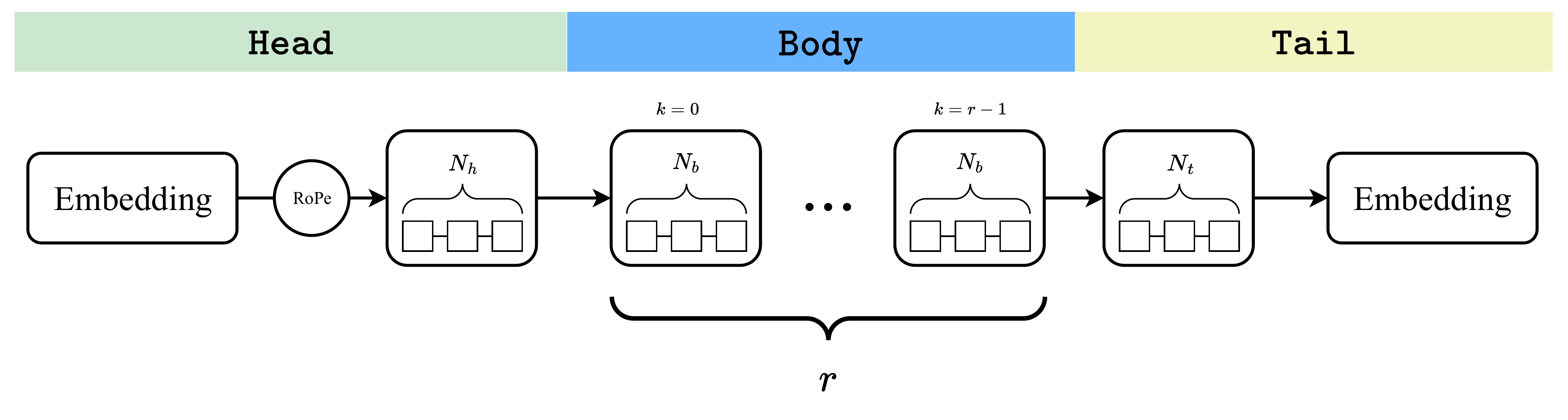}
    \caption{Architecture of \abe method. The model consists of three main components: \texttt{Head} (green), \texttt{Body} (blue), and \texttt{Tail} (yellow). Input embeddings first pass through the \texttt{Head}, where they are converted from token space to concept space. The embeddings are then iterated over $r$ times $(k=0, 1, ..., r-1)$ by the \texttt{Body}. The final embeddings are then processed by the \texttt{Tail} and projected back into token space. $N_h$,$N_b$ and $N_t$ show the number of Transformers blocks in each component.}
    \vspace{-1em}
    \label{fig:abbie}
\end{figure}
\abe consists of three structural groups and follows the structure of a decoder-only architecture \citep{DBLP:conf/nips/VaswaniSPUJGKP17, DBLP:journals/corr/abs-2412-19437}. We term the three groups \texttt{Head}, \texttt{Body}, and \texttt{Tail}. This separation is similar to the structure presented by \citet{geiping2025scalingtesttimecomputelatent}, but unique in that \abe does not require additional projection matrices or large training iterations.  A visual representation of the separation is shown in Fig. \ref{fig:abbie}. The primary distinction between \abe and a standard Transformer lies in the iterative structure of the \texttt{Body}: instead of processing the tokens once, \abe applies the body stack repeatedly across iterations. In \abe, each structural group contains $N_h$, $N_b$, $N_t$ Transformer blocks respectively. The \texttt{Body} further operates recursively over $r$ iterations. This structural grouping can be seen as a generalization of the original Transformer architecture, which emerges as a special case under our formulation when the iteration number $r = 1$. The \texttt{Head} contains the embedding layer $W$ and $N_h$ Transformer blocks. Its role is to transforms the embedded token representations into concept space through the detokenization process \citep{DBLP:journals/corr/abs-2410-05864}. Concept space, is an internal space in the Transformer body, described by \citet{DBLP:journals/corr/abs-2410-05864,DBLP:journals/corr/abs-2412-06769}. Once in concept space, the \texttt{Body} iterates over the residual stream using one of two operational variants: \co or \diff. In \co, during each iteration, the \texttt{Body} operates directly on the output of the previous iteration. In \diff, the outputs of each iteration are added to the original input, forming a residual chain. Finally, the \texttt{Tail} maps the representations back to token space before unembedding. 

\subsection{Operation}\label{sec:operation}
\begin{figure}[htbp]
    \centering
    \includegraphics[width=\textwidth]{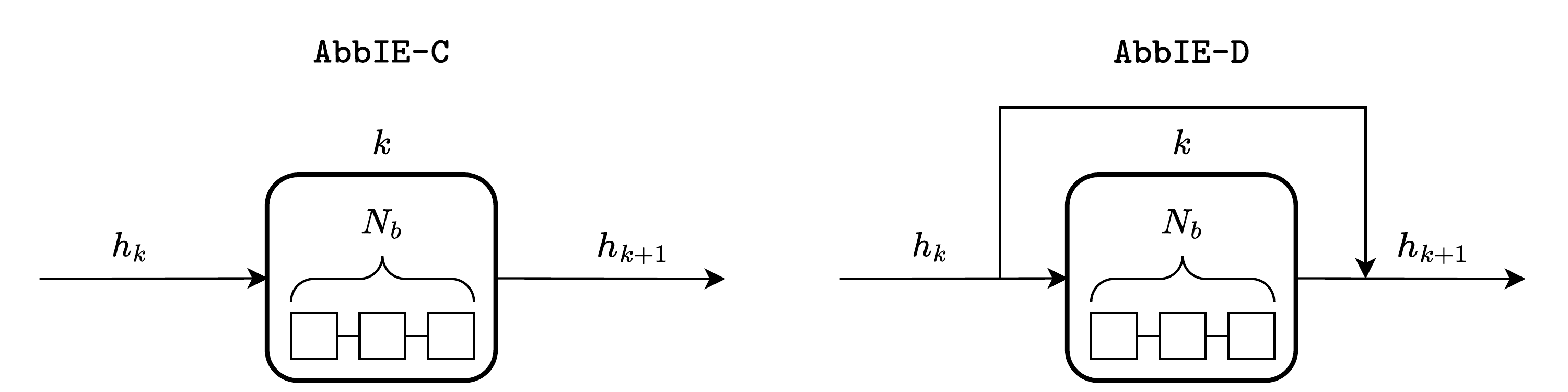}
    \caption{Architectural comparison of the two \abe variants. \co (left) uses the Transformer's existing residual connection as means of input injection, whereas \diff (right) introduces and additional inter-iteration residual. Although similar it is not possible, given an arbitrary \diff instance, to produce an \co instance that computes the same function.}
    \label{fig:abbie_struc}
\end{figure}

We adopt a recurrent formulation to allow the model to spend a variable amount of computation per token. Prior work has demonstrated that allowing Transformers to reprocess their residual streams across multiple passes improves performance \citep{DBLP:journals/corr/abs-2112-00114, DBLP:journals/corr/abs-2412-06769}. \abe has two operational variants: a continuous chain-of-thought inspired mode (\co) and a diffusion-inspired mode (\diff). The methods differ in the way they use the residual stream between iterations in the \texttt{Body}. We motivate the creation of both methods through the Path Independence (PI) property \citep{DBLP:conf/nips/AnilPLTWBKG22}. PI, introduced by \citet{DBLP:conf/nips/AnilPLTWBKG22}, is a relaxed form of a contraction mapping (lacks strictly decreasing requirement). Formally, a function $f$ is PI if repeated application it to an input $x$ leads to a fixed point $z^*$:
\begin{equation}
f^{(\infty)}(x,z_0)=z^*, \text{for any}~ z_0
\end{equation}
where $z_0$ is some initial state and the superscript $^{(\cdot)}$ indicates repeated application. PI has been shown to correlate with a model's ability to generalize across iteration counts, i.e. \textbf{upwards generalization}. While \citet{DBLP:conf/nips/AnilPLTWBKG22} identify the necessary components for PI (dependence on the original input at each iteration and the injection of noise in the first iteration), these do not preclude other methods from achieving fixed points and thus upwards generalization. In this work, we omit $z_0$ or implicitly set $z_0 = \mathbf{0}$. Therefore we can use a formulation which only relies on the input as:
\begin{equation}
    f^{(n)}(x_n) \approx f^{(n+1)}(x_{n+1}) ~\text{as}~ n \to \infty
\end{equation}
where the subscript shows the iterant index. Therefore, we are only concerned with the inclusion of inputs into each iteration. Transformers inherently depend on the original inputs via the residual stream, suggesting that simply looping within concept space may suffice for generalization. Based on this idea, we define our first variant, \co, as direct recursion over the residual stream (Fig.~\ref{fig:abbie_struc}), akin to continuous chain-of-thought models \citep{DBLP:journals/corr/abs-2412-06769}. Formally, we can define \co as:
\begin{equation}\label{co_update}
    h_{k+1} = B(h_k), \quad \text{for } k = 0, \ldots, N - 1
\end{equation}
where $B(\cdot)$ is the \texttt{Body} component. The second variant, \diff, re-injects inputs into each block through an inter-iteration residual stream. Formally, this can be defined as:
\begin{equation}\label{diff_update}
    h_{k+1} = B(h_k) + h_k, \quad \text{for } k = 0, \ldots, N - 1
\end{equation}
Since each of the Transformer blocks in the \texttt{Body} has a residual connection, \co and \diff could seem equivalent. The additional residual connection around the whole \texttt{Body} block, however, reinforces the relative strength of the information contained in the input embeddings $h_0$ compared to the cumulative updates originating from the Attention and FFN components in the Transformer blocks. To see this, we can define the delta between iteration states induced by $B(\cdot)$ as:
\begin{equation}
\Delta h_k =\sum_{i=0}^{N_b - 1} (\text{Attn}_i + \text{FFN}_i)
\end{equation}
\noindent
where  each element in the sum depends on all the previous elements. Then, for \co, from \eqref{co_update}, the subsequent iteration output becomes $h_{k+1} = h_k + \Delta h_k$ while for \diff, from \eqref{diff_update}, it is $h_{k+1} = 2 h_k + \Delta h_k$. To achieve the same output composition, the model should learn to make $h_k[\diff] = h_k[\co] / 2$, but then this would make $\Delta h_k[\diff]$ different from $\Delta h_k[\co]$ in an unpredictable, non-linear way.

\vspace{-1.5em}
\section{Experimental Setup}
In this section we describe the training procedures used throughout this work, along with the baselines and evaluations used to assess model performance.

\subsection{Baselines}
To evaluate the effectiveness of \abe, we conduct comparisons against both standard and recursive Transformer baselines. \depth is our implementation of the recursive method proposed by \citet{geiping2025scalingtesttimecomputelatent}. Like \abe, \depth applies iteration over a shared set of Transformer blocks in latent space. However, instead of relying on the residual stream for input re-injection, \depth concatenates the input embeddings to, initially a randomly sampled vector, and in subsequent iterations to the output of the previous iteration. This concatenated vector is projected back to the model dimension. While the original method uses non-standard width-to-depth ratios, we use GPT-2-like architectural proportions to ensure fair comparison. Although \citet{geiping2025scalingtesttimecomputelatent} propose a custom sandwich normalization strategy, they report that at small scales, most normalization approaches (pre-norm, post-norm, etc.) perform comparably. Accordingly, we adopt a consistent pre-norm configuration across all baselines.

For our standard Transformer baseline (\texttt{Std}), we use a decoder-only architecture loosely inspired by the design choices in \citet{grattafiori2024llama3herdmodels}, with proportions similar to those used in GPT-2 \citep{radford2019language}. This model lacks any form of iteration and serves as the canonical single-pass baseline.
\subsection{Training Configuration}
All models, defined in Table~\ref{model_config}, are trained using standard Transformer training practices. We define the compute-optimal token budget (COT) as \textbf{20 tokens per model parameter}, consistent with established empirical scaling laws \citep{DBLP:journals/corr/abs-2203-15556}. All models are trained up to the COT budget, unless specified otherwise. In specific analyses involving perplexity scaling, we include training runs beyond the COT budget, but all baseline comparisons are made at the compute-optimal point. Training is performed using the AdamW optimizer with $\beta_1 = 0.9$ and $\beta_2 = 0.95$. We use two global batch sizes: 256 samples for the small-batch (SB) regime and 1024 samples for the large-batch (LB) regime. The two regimes allows us to test how recurrent processing interacts with optimization dynamics under different gradient noise conditions \cite{mccandlish2018empiricalmodellargebatchtraining}. All models are trained on fixed-length sequences of 2048 tokens.
We adopt a Warmup-Stable-Decay learning rate schedule. We use linear warmup lasting $0.25 \times \text{COT}$ steps. We use a cosine decay over $0.20 \times \text{COT}$ steps, in line with best practices from \citet{DBLP:journals/corr/abs-2412-19437}. Tied input-output embedding matrices are used across all models, ensuring that observed improvements stem from the model body itself rather than bi-gram memorization or shortcut learning. Architecturally, we adopt a modern Transformer design with Grouped Query Attention (GQA) \citep{ainslie-etal-2023-gqa, grattafiori2024llama3herdmodels}, Rotary Positional Embeddings (RoPE) \citep{10.1016/j.neucom.2023.127063}, SiLU activation \citep{elfwing2017sigmoidweightedlinearunitsneural}, Pre-norm residual structure \citep{DBLP:conf/icml/XiongYHZZXZLWL20} and RMSNorm\citep{zhang2019rootmeansquarelayer}.

\begin{table}
  \caption{Model configurations for evaluations across two model sizes and two batch sizes. We use GPT-2-like architectures. $N_{h,b,t}$ show component numbers, $d_{model}$ and FFN size define the width of the network. Attn heads and KV heads define the GQA paramters and RoPE $\theta$ shows the base used in the rotary positional embeddings.}
  \label{model_config}
  \centering
  \begin{tabular}{cccccccccc}
    \toprule
    Size & BS & $N_h$ & $N_b$ & $N_t$ & $d_{model}$ & FFN size & Attn heads & KV heads & RoPE $\theta$\\
    \midrule
    200M & 256  & 2&10&2&  1024 &4096&16&8&10,000  \\
    200M & 1024  & 2&10&2& 1024  &4096& 16&8&10,000 \\
    350M & 1024  & 2&20&2 &1024   &4096& 16&8&10,000\\
    \bottomrule
  \end{tabular}
  \vspace{-1.4em}
\end{table}%
For the smaller size models we have performed 5 training runs each with different seeds to verify the statistical significance of our experimentation; however for the large models we have only conducted a single training run. \textit{Best performance is reported}. No hyperparameter tuning is performed per variant. All experiments were run on a single H100 GPU.

\subsection{Training Dataset and Evaluation Metrics}
We train on a mixture of datasets proposed by \citet{vonplaten2023smollm} which provides models with a high-quality mix to achieve a high in-context learning performance. We use 80\% of FineWeb-Edu \cite{penedo2024finewebdatasetsdecantingweb}, 10\% Cosmopedia-v2 \cite{benallal2024cosmopedia} and 10\% Python-Edu \cite{Kocetkov2022TheStack}.
FineWeb-Edu is a filtered subset of the broader FineWeb dataset, curated to emphasize high-quality, diverse, and educationally valuable text. Cosmopedia-v2 consists of synthetic encyclopedic content generated to enhance factual recall and structured knowledge. Python-Edu contains educational programming content, with a focus on natural language explanations of Python code and logic.

For evaluation, we focus on a suite of in-context learning (ICL) benchmarks particularly in the domains of commonsense reasoning and language understanding. These benchmarks are: \textbf{HellaSwag} \citep{zellers2019hellaswagmachinereallyfinish}, \textbf{LAMBADA} \citep{paperno2016lambadadatasetwordprediction}, \textbf{ARC-Easy} \citep{allenai_arc} and \textbf{CommonsenseQA} \citep{talmor2019commonsenseqaquestionansweringchallenge}.
At the model sizes that we examine, some ICL tasks are unsuitable for evaluation as they do not improve above the random baseline level. One of these tasks is CommonsenseQA which we intentionally include to show that iteration scaling significantly improves our reasoning capabilities. Each evaluation is performed in the zero-shot setting. We report accuracy as the primary metric across all tasks. Our goal in selecting these benchmarks is to assess effectiveness of iterative refinement mechanisms in reasoning-centric tasks that benefit from multi-step internal processing \cite{biran2024hoppinglateexploringlimitations}.

\vspace{-1em}
\section{Results}
\vspace{-1em}
In this section we present the results from evaluating the \abe method, comparing it against current approaches in recurrent and standard Transformers. With these experiments we answer the following research questions:

\textbf{RQ1}: Can \abe achieve a fixed point iteration by using the residual as input injection? (Sec.~\ref{rq1})

\textbf{RQ2}: Can \abe reach a lower perplexity for the same size and token budget? (Sec.~\ref{rq2})

\textbf{RQ3}: Can \abe achieve upward generalization with only two training iterations? (Sec.~\ref{rq3})

\textbf{RQ4}: Can \abe's scale test time performance? (Sec.~\ref{rq3})

\subsection{Fixed Point (RQ1)}\label{rq1}
\begin{figure}[t]
    \centering
    \includegraphics[width=\linewidth]{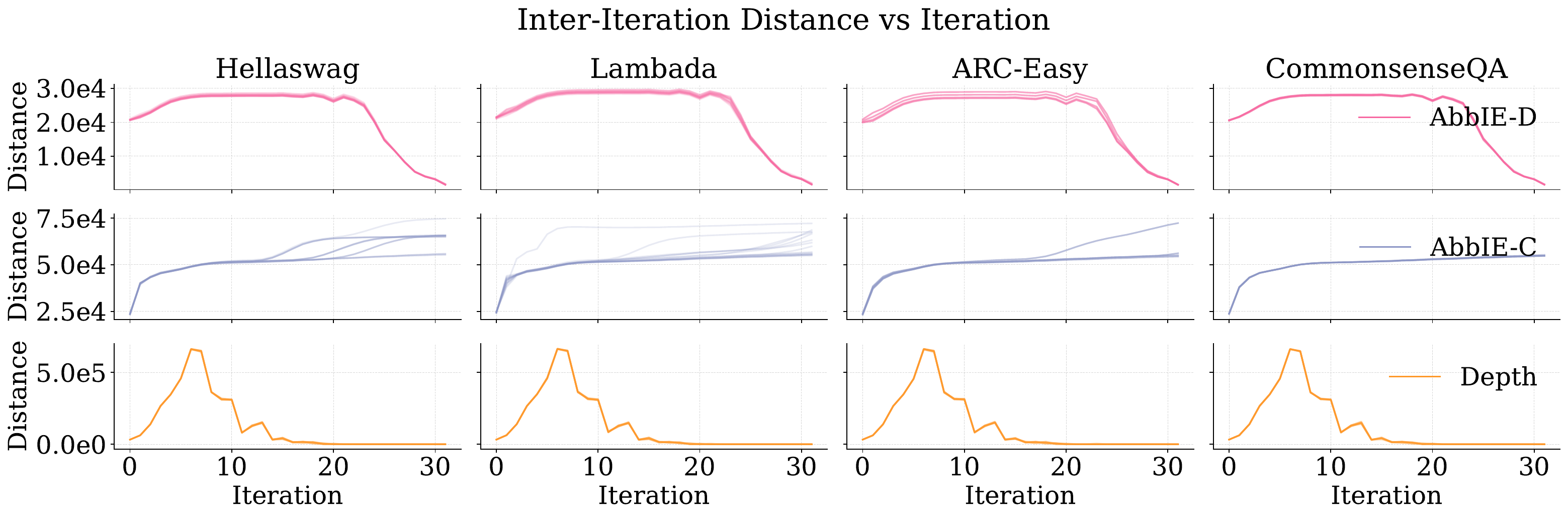}
    \caption{Inter-iteration distances vs number of iterations for 16 random samples from various benchmarks (Hellaswag, Lambada, ARC-Easy, and CommonsenseQA). We see that \diff (top) and \depth (bottom) reach a fixed point. This implies greater stability and better upward generalization capabilities. \co (middle) consistently diverges.}
    \label{fig:distance}
    \vspace{-1em}
\end{figure}

To evaluate whether \abe converges to a fixed point in iteration count, we measure the distance between the input and output embeddings at each iteration. This analysis is performed on randomly sampled data points from the ICL evaluation sets. As a control, we include the \depth\ method, which is known to be path independent by construction and thus guaranteed to reach a fixed point. Figure~\ref{fig:distance} summarizes the results. We observe that \diff exhibits smooth decay past a certain iteration, eventually converging. This behavior is shared with the \depth method showing that \diff, in the samples we have evaluated on, reaches a fixed point. In contrast, \co fails to converge after 32 iterations. This shows that the residual steam, present in the body of Transformer, is in itself not sufficient and an extra inter-iteration residual connection is needed. Due to \co inability to converge it is excluded from further analyses.

\subsection{Performance scaling and computational expenditure (RQ2)}\label{rq2}
\begin{figure}[ht]
    \centering
    \begin{subfigure}[t]{0.48\textwidth}
        \centering
        \includegraphics[width=\textwidth]{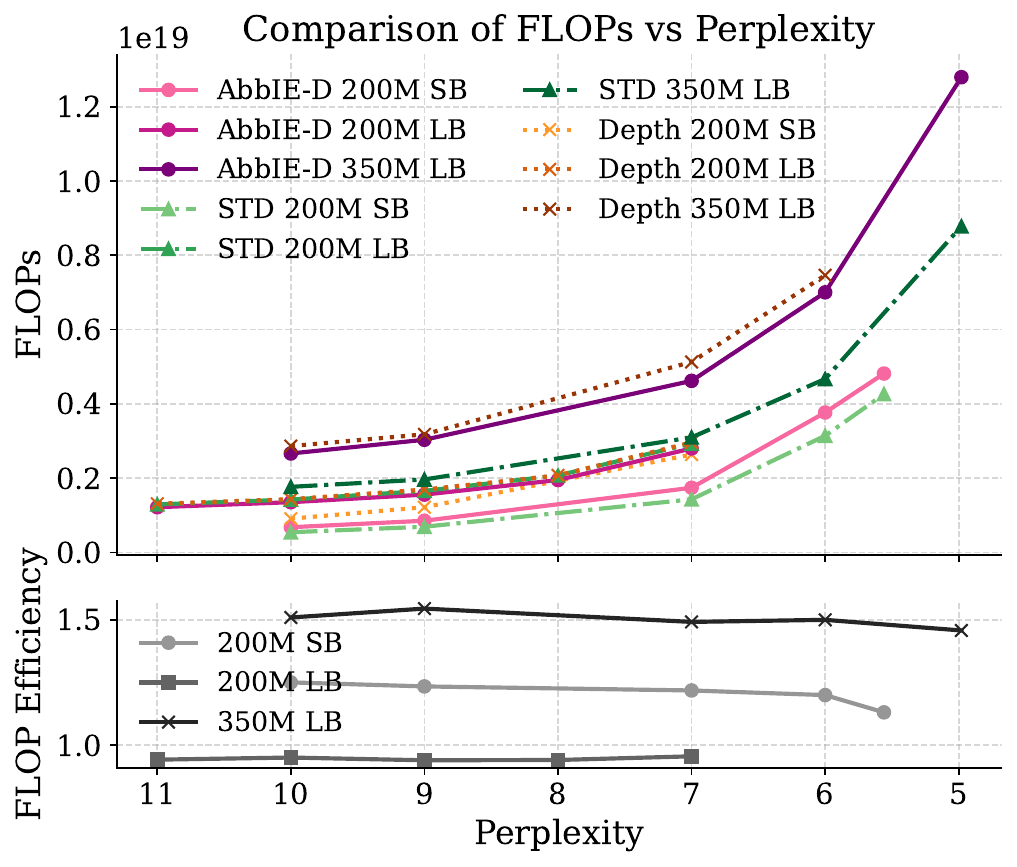}
        \caption{Computational expenditure of \abe and its scaling with perplexity (Top). FLOP efficiency (Bottom), defined as the ratio of FLOPs spent on \abe to FLOPs spent on \std, has a negative gradient. This shows that with long training runs, the FLOP efficiency will improve.}
        \label{fig:equiperplexity}
    \end{subfigure}
    \hfill
    \begin{subfigure}[t]{0.48\textwidth}
        \centering
        \includegraphics[width=\textwidth]{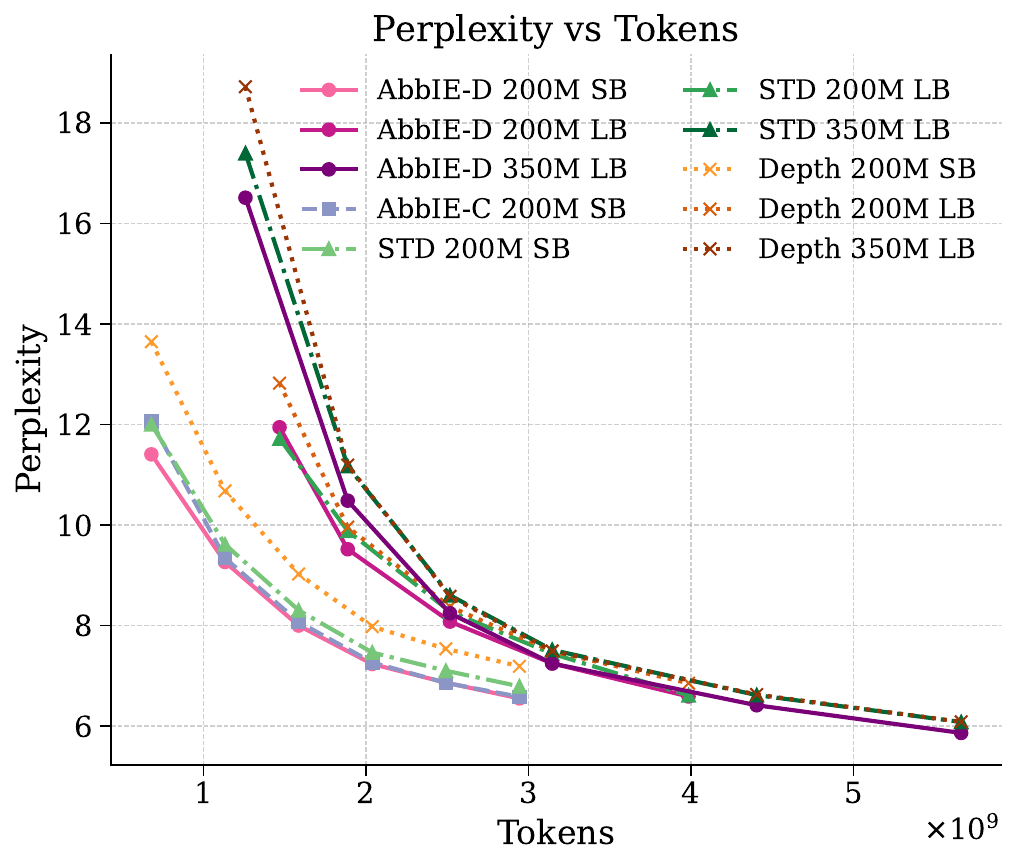}
        \caption{Perplexity vs token count follows standard scaling law patterns, with \abe having a roughly 5\% better perplexity than \std. \depth follows the same trend, but with a worse perplexity.}
        \label{fig:equi-token}
    \end{subfigure}
    \caption{Performance scaling and computational expenditure of \abe vs \std vs \depth. The figures shows that \abe follows typical scaling laws and has an improved perplexity. For token amounts near COT, the computational cost of \abe is typically higher than that of \std but lower than \depth.}
    \label{fig:mainfigure}
\end{figure}

To evaluate the performance of \abe we show how perplexity evolves as a function of token budget. Figure~\ref{fig:equi-token} summarizes this comparison.
Across all configurations, \abe consistently outperforms both the standard Transformer and \depth baseline in terms of perplexity.

\abe follows scaling trends consistent with standard Transformers, showing steady reductions in perplexity with increased model size and token count. This indicates that \abe retains the favorable scaling properties of Transformer architectures while offering improved perplexity.

Next, we evaluate the relationship between computational cost and model performance during training. As other recurrent Transformers require many iterations to train, we find it important to demonstrate that the training costs of \abe are much smaller than a typical recurrent formulation and are comparable with standard Transformers. Following the methodology from \citet{DBLP:journals/corr/abs-2203-15556}, we estimate the floating point operations (FLOPs) required to reach various perplexity thresholds. Results are shown in Figure~\ref{fig:equiperplexity}. \abe follows the same scaling laws as standard Transformers: lower perplexities require more computation. However, it generally incurs a higher FLOP cost. Despite this, the additional cost decreases over time. To show this we define FLOP efficiency as the ratio of \abe FLOPs to standard Transformer FLOPs. The gradient of FLOP efficiency in Figure~\ref{fig:equiperplexity} is negative. Therefore, for very long training runs, on the order of $20\times$ to $200\times$ the compute-optimal budget, which is common in large-scale training, we hypothesize the total computational cost of \abe will converge to that of the standard Transformer. This is because \abe maintains a consistent perplexity advantage, while the marginal improvements in perplexity become harder to achieve as training progresses. Crucially we show that \diff's improvements in perplexity and scaling stability are valid across model and batch sizes.
\subsection{Test time scaling (RQ3, RQ4)}\label{rq3}

\begin{figure}[]
    \centering
    \begin{subfigure}[t]{0.48\textwidth}
        \centering
        \includegraphics[width=\textwidth]{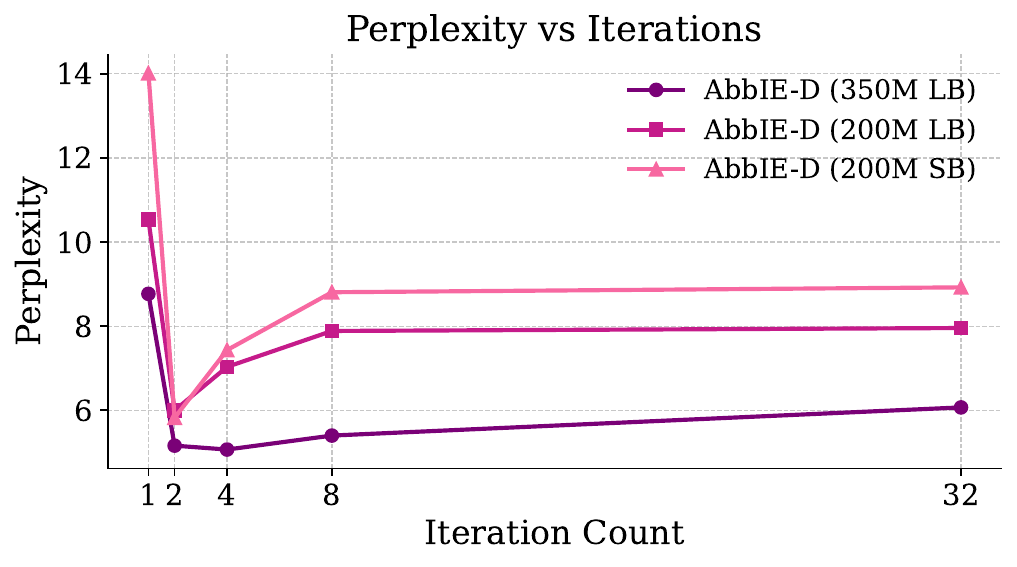}
        \caption{Comparison of perplexity across iterations for \diff models of different sizes. The 350M parameter model continues to improve perplexity past the number of trained iterations. This shows that it has generalized in iteration count.}
        \label{fig:perpA}
    \end{subfigure}
    \hfill
    \begin{subfigure}[t]{0.48\textwidth}
        \centering
        \includegraphics[width=\textwidth]{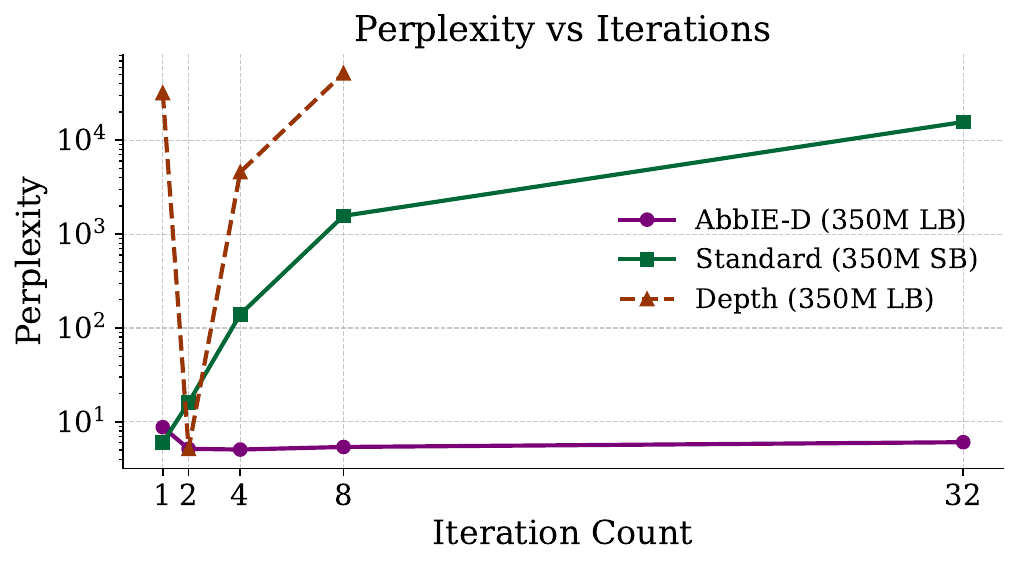}
        \caption{Comparison of perplexity (log scale) across iterations for \diff, \depth and \std. \diff exhibits stable perplexity, whereas \depth experiences perplexity collapse at $r\neq2$.  \std is included to show the behavior of a non-iterative model in a iterative setting.}
        \label{fig:perpB}
    \end{subfigure}
    \caption{Perplexity trends across iteration counts for different model variants. Sub figure (a) shows a detailed comparisons between \diff configurations. }
    \label{fig:perpScale}
\end{figure}

To evaluate the scaling characteristics of \diff we measure the perplexity at different iteration counts. We compare versus the \depth baseline. Figure~\ref{fig:perpA} shows that \diff possesses a stable perplexity in the number of iterations. The 200M parameter versions achieve their lowest perplexity at $r = 2$, which matches their training regime. However, the 350M parameter version achieves its lowest perplexity at $r = 4$, twice the training regime. This suggests that 
\diff at the larger size was able to successfully upward generalize (\textbf{RQ3}). This finding is also confirmed by the ICL task performance. In contrast, Figure~\ref{fig:perpB} shows that \depth cannot generalize in the same training regime. 

Figure~\ref{fig:ICLScaling} shows the ICL task scaling of \diff compared to \std and \depth(only at $r=2$, where perplexity is stable). We show that \diff's perplexity scaling has translated into improved ICL task performance. Even when the perplexity at $r = 8$ is higher than the minimum the ICL task performance continues to improve. This is important, as most recurrent Transformers's performance plateaus at the number of training iterations. Conversely, we see that once generalization is achieved in \diff, even though perplexity begins to increase after a certain point, ICL task performance continues to improve. This also shows that the relationship between ICL and perplexity is not strict.\textbf{ Notably we show that \diff continues to improve performance at $4 \times$ the training iteration count, which no other general recursive transformer has shown.} (\textbf{RQ4})

For \diff 350M we see improvements of up to 12\% on HellaSwag. In CommonsenseQA at 8 iterations we see a performance significantly higher than the random baseline which was achieved by all other models. However, as no other method was able to surpass the random baseline, we cannot claim direct percentage improvement in CommonsenseQA. However, \textbf{we show that \abe improves the emergent reasoning capabilities of the Transformer.} \textbf{The magnitude of the ICL task performance is on par with other models of a similar size trained to COT} \cite{amd2023amdllama135m,vonplaten2023smollm}. The rest
of the comparisons are summarized in Table \ref{tab:full_res}. 
For \diff 200M we closely match the performance of the standard Transformer with a distinctive reduction of performance at $r=1$. However, this difference disappears for \diff 350M, which suggest some critical model size larger than 200M has to be used for the model to have the spare capacity to learn to improve the residual stream with iterations. The performance gains also remain across different batch sizes and therefore gradient noise regimes. This indicates that the \abe approach is robust to the sharp minima present in larger batch training.

\begin{table}[h]
\centering
\caption{Zero-shot performance (\%) across recursive iterations on standard In-Context Learning (ICL) benchmarks. We report the best result from each method-category. \diff consistently outperforms all other approaches and is the only method to achieve a CSQA score (For the Commensense Question Answering task) above the random baseline. From the results, we can see \depth achieves $29.6$ for 200M model. But by comparing it against Fig.~\ref{fig:perpB} we can observe that the model's perplexity has collapsed (indicating model collapse). Therefore, the accuracy of $29.6$ potentially is a random outcome. }
\label{tab:full_res}
\begin{tabular}{lllcccc}
\toprule
\textbf{Size} & \textbf{BS} & \textbf{Model} & \textbf{Hella\-Swag} & \textbf{LAMBADA} & \textbf{ARC-E} & \textbf{CSQA} \\
\midrule
\multirow{18}{*}{\textbf{200M}} & \multirow{9}{*}{\textbf{256}} 
  & \std           & 30.1 & \textbf{24.2} & 45.6 & 20.0 \\
& & \diff ($r=1$)  & 30.2 & 6.0  & 42.5 & 20.0 \\
& & \diff ($r=2$)   & \textbf{31.4} & 23.1 & 45.9 & 20.0 \\
& & \diff ($r=4$)   & 31.3 & \textbf{24.2} & 48.7 & 20.0 \\
& & \diff ($r=8$)   & 31.3 & 22.7 & \textbf{49.2} & 20.0 \\
& & \depth ($r=1$)  & 25.0 & 0.0  & 26.8 & 20.0 \\
& & \depth ($r=2$)  & 29.7 & 22.2 & 46.3 & 20.0 \\
& & \depth ($r=4$)  & 25.0 & 0.0  & 27.5 & 29.6 \\
& & \depth ($r=8$)  & 25.0 & 0.0  & 25.0 & 20.0 \\
\cmidrule(lr){2-7}

& \multirow{9}{*}{\textbf{1024}} 
  & \std           & 30.2 & 23.2 & 47.0 & 20.0 \\
& & \diff ($r=1$)  & 29.2 & 11.0 & 41.8 & 20.0 \\
& & \diff ($r=2$)   & 30.0 & 21.8 & 45.6 & 20.0 \\
& & \diff ($r=4$)   & 30.4 & 25.1 & 46.5 & 20.0 \\
& & \diff ($r=8$)   & \textbf{30.6} & \textbf{25.9} & \textbf{47.3} & 20.0 \\
& & \depth ($r=1$)  & 25.0 & 0.0  & 26.5 & 20.0 \\
& & \depth ($r=2$)  & 29.4 & 22.5 & 45.6 & 20.0 \\
& & \depth ($r=4$)  & 25.0 & 0.0  & 28.5 & 29.6 \\
& & \depth ($r=8$)  & 25.0 & 0.0  & 25.8 & 20.0 \\
\midrule

\multirow{9}{*}{\textbf{350M}} & \multirow{9}{*}{\textbf{1024}} 
  & \std            & 32.5 & 26.0 & 48.2 & 20.0 \\
& & \diff ($r=1$)   & 31.9 & 25.0 & 48.5 & 20.0 \\
& & \diff ($r=2$)   & 33.6 & 27.8 & 49.0 & 21.0 \\
& & \diff ($r=4$)   & 34.5 & 28.0 & 50.8 & 25.0 \\
& & \diff ($r=8$)   & \textbf{34.8} & \textbf{28.0} & \textbf{51.3} & \textbf{27.6} \\
& & \depth ($r=1$)  & 25.0 & 0.0  & 25.4 & 20.0 \\
& & \depth ($r=2$)  & 32.5 & 26.5 & 49.3 & 20.0 \\
& & \depth ($r=4$)  & 25.0 & 0.0  & 28.5 & 20.0 \\
& & \depth ($r=8$)  & 25.0 & 0.0  & 26.4 & 20.0 \\
\midrule
\multicolumn{3}{c}{\textbf{Random Baseline}} & \textbf{25.0} & \textbf{0.0} & \textbf{25.0} & \textbf{20.0} \\
\bottomrule
\end{tabular}
\vspace{-0.5em}
\end{table}

\section{Background and Related Work}

\textbf{Latent space}
Transformers perform the majority of their computation in a high-dimensional latent space. While the architecture of a standard Transformer is uniform across layers, the representations it produces are not. Recent work has shown that Transformer latent space can be broadly divided into two distinct regions \cite{DBLP:journals/corr/abs-2410-05864, DBLP:journals/corr/abs-2412-06769}. The first, which we refer to as \textit{token space} consists of the low-level hidden states produced by the embedding matrix, where each vector corresponds to a subword unit generated by the tokenizer. The second, which we refer to as the \textit{concept space}, consists of higher-level semantic representations, which evolve through the depth of the Transformer \cite{DBLP:conf/emnlp/KatzB23}. These are ultimately mapped back into token space by the final Transformer blocks and the output (enembedding) layer \citep{DBLP:journals/corr/abs-2410-05864}.
The transition from token space to concept space, referred to as \textit{detokenization} \citep{DBLP:journals/corr/abs-2410-05864, elhage2022solu, DBLP:journals/tmlr/GurneeNPHTB23} occurs within the early layers of the model. Within concept space, several studies have demonstrated that Transformer blocks can be reordered, skipped, or reused \citep{fan2019reducing, DBLP:conf/acl/ElhoushiSLHWL0A24, shen2025lunar} without significantly impacting performance. This suggests that individual layers specialize in specific conceptual transformations, rather than relying on rigid positional dependencies.

\textbf{Path independence}
Path independence, introduced by \citet{DBLP:conf/nips/AnilPLTWBKG22}, is a property of an iterative model to converge to a fixed point regardless of its initialization. The authors argue that PI strongly correlates with high accuracy on out-of-distribution samples (upward generalization). Intuitively, convergence to a fixed point reflects stable refinement of representations; in contrast, non-convergent models may exhibit erratic updates that harm performance. Formally, a method is path independent if it converges to the same limiting behavior from any initial state \cite{DBLP:conf/nips/AnilPLTWBKG22}. In this work, we use concepts from path independence (input injection) and  empirically show the existence of a fixed point by analyzing the sequence of representations across iterations.

\textbf{Recurrent Transformers}
Recurrent formulations of Transformer networks have gained attention as a promising approach to overcome the scaling limitations of standard architectures. They benefit from increased computation depth and improve performance on certain tasks \cite{geiping2025scalingtesttimecomputelatent}. These methods typically use weight tying across layers; some, such as the Universal Transformer \citep{DBLP:conf/iclr/DehghaniGVUK19}, also include depth embeddings, which can limit scalability at test time. Others incorporate inter-iteration projections to sustain performance at increasing depths. 
A common approaches in recurrent Transformers is the re-injection of inputs at each iteration, often coupled with random noise \citep{geiping2025scalingtesttimecomputelatent, DBLP:conf/iclr/Yang0NP24, DBLP:conf/icml/GiannouRS0LP23} to aid in the performance and extrapolation capabilities of the model. The method proposed by \citet{geiping2025scalingtesttimecomputelatent} shares architectural similarities with \abe but differs in several key aspects. Their model initializes the latent space with random vectors and is both trained over a range of randomly sampled iteration count. As a result, its performance saturates near the average number of training iterations.  In contrasts, \abe leverages the internal structure of the Transformer's latent space and residual connections to generalize across arbitrary iteration counts without the need for specialized embeddings or projections. Additionally, \abe is trained with just two iterations and achieves stable perplexity at test time regardless of depth, offering a more efficient and generalizable approach.

\section{Discussion and Limitations}
Recurrent Transformer architectures have demonstrated strong performance on a range of downstream tasks, outperforming non-recurrent models. However, these methods are either highly modified versions of the Transformer network, which are unsuitable for general language modeling or have not been evaluated in a controlled, one-to-one comparison against identical standard Transformer architectures. As a result, it remains unclear how much of the observed gains are due to recurrence itself versus other architectural or training differences. A major limitation of existing recurrent models is their consistently larger computational cost. They require multiple iterations per token and lack the ability to adaptively reduce computation when recurrence is unnecessary. This restricts their applicability in resource-constrained settings and reduces throughput. In this work, we introduced a new approach, \abe, that achieves upwards generalization with as few as two iterations. Although lower than all other recurrent Transformers, for token budgets close to COT, \abe still uses more FLOPs. Our method produces lower perplexity with increasing loop count, enabling dynamic control over compute at inference time. Our experimental results show that \abe reaches a \textbf{5\%} improvement in perplexity and a upward of \textbf{12\%} improvement in ICL tasks with improved emergent properties. When run with a single iteration, our model matches the perplexity of a standard Transformer. We demonstrate that the representations in our method gradually converge toward a fixed point. However, prior work suggests that Transformer dynamics may not align perfectly with fixed-point behavior. This warrants further investigation into the nature of iterative refinement in attention-based models. Future work will explore enhanced training objectives and normalization strategies to further improve perplexity and convergence.

\bibliography{main}

\begin{thebibliography}{52}
\providecommand{\natexlab}[1]{#1}
\providecommand{\url}[1]{\texttt{#1}}
\expandafter\ifx\csname urlstyle\endcsname\relax
  \providecommand{\doi}[1]{doi: #1}\else
  \providecommand{\doi}{doi: \begingroup \urlstyle{rm}\Url}\fi

\bibitem[Ainslie et~al.(2023)Ainslie, Lee-Thorp, de~Jong, Zemlyanskiy, Lebron, and Sanghai]{ainslie-etal-2023-gqa}
J.~Ainslie, J.~Lee-Thorp, M.~de~Jong, Y.~Zemlyanskiy, F.~Lebron, and S.~Sanghai.
\newblock {GQA}: Training generalized multi-query transformer models from multi-head checkpoints.
\newblock In H.~Bouamor, J.~Pino, and K.~Bali, editors, \emph{Proceedings of the 2023 Conference on Empirical Methods in Natural Language Processing}, pages 4895--4901, Singapore, Dec. 2023. Association for Computational Linguistics.
\newblock \doi{10.18653/v1/2023.emnlp-main.298}.
\newblock URL \url{https://aclanthology.org/2023.emnlp-main.298/}.

\bibitem[Allal et~al.(2025{\natexlab{a}})Allal, Lozhkov, Bakouch, Bl{\'{a}}zquez, Penedo, Tunstall, Marafioti, Kydl{\'{\i}}cek, Lajar{\'{\i}}n, Srivastav, Lochner, Fahlgren, Nguyen, Fourrier, Burtenshaw, Larcher, Zhao, Zakka, Morlon, Raffel, von Werra, and Wolf]{DBLP:journals/corr/abs-2502-02737}
L.~B. Allal, A.~Lozhkov, E.~Bakouch, G.~M. Bl{\'{a}}zquez, G.~Penedo, L.~Tunstall, A.~Marafioti, H.~Kydl{\'{\i}}cek, A.~P. Lajar{\'{\i}}n, V.~Srivastav, J.~Lochner, C.~Fahlgren, X.~Nguyen, C.~Fourrier, B.~Burtenshaw, H.~Larcher, H.~Zhao, C.~Zakka, M.~Morlon, C.~Raffel, L.~von Werra, and T.~Wolf.
\newblock Smollm2: When smol goes big - data-centric training of a small language model.
\newblock \emph{CoRR}, abs/2502.02737, 2025{\natexlab{a}}.
\newblock \doi{10.48550/ARXIV.2502.02737}.
\newblock URL \url{https://doi.org/10.48550/arXiv.2502.02737}.

\bibitem[Allal et~al.(2025{\natexlab{b}})Allal, Lozhkov, Bakouch, Blázquez, Penedo, Tunstall, Marafioti, Kydlíček, Lajarín, Srivastav, Lochner, Fahlgren, Nguyen, Fourrier, Burtenshaw, Larcher, Zhao, Zakka, Morlon, Raffel, von Werra, and Wolf]{allal2025smollm2smolgoesbig}
L.~B. Allal, A.~Lozhkov, E.~Bakouch, G.~M. Blázquez, G.~Penedo, L.~Tunstall, A.~Marafioti, H.~Kydlíček, A.~P. Lajarín, V.~Srivastav, J.~Lochner, C.~Fahlgren, X.-S. Nguyen, C.~Fourrier, B.~Burtenshaw, H.~Larcher, H.~Zhao, C.~Zakka, M.~Morlon, C.~Raffel, L.~von Werra, and T.~Wolf.
\newblock Smollm2: When smol goes big -- data-centric training of a small language model, 2025{\natexlab{b}}.
\newblock URL \url{https://arxiv.org/abs/2502.02737}.

\bibitem[{AMD}(2023)]{amd2023amdllama135m}
{AMD}.
\newblock Amd-llama-135m: A 135m parameter language model trained on amd instinct mi250 accelerators.
\newblock \url{https://huggingface.co/amd/AMD-Llama-135m}, 2023.
\newblock Accessed: 2025-05-15.

\bibitem[Anil et~al.(2022)Anil, Pokle, Liang, Treutlein, Wu, Bai, Kolter, and Grosse]{DBLP:conf/nips/AnilPLTWBKG22}
C.~Anil, A.~Pokle, K.~Liang, J.~Treutlein, Y.~Wu, S.~Bai, J.~Z. Kolter, and R.~B. Grosse.
\newblock Path independent equilibrium models can better exploit test-time computation.
\newblock In S.~Koyejo, S.~Mohamed, A.~Agarwal, D.~Belgrave, K.~Cho, and A.~Oh, editors, \emph{Advances in Neural Information Processing Systems 35: Annual Conference on Neural Information Processing Systems 2022, NeurIPS 2022, New Orleans, LA, USA, November 28 - December 9, 2022}, 2022.
\newblock URL \url{http://papers.nips.cc/paper\_files/paper/2022/hash/331c41353b053683e17f7c88a797701d-Abstract-Conference.html}.

\bibitem[Bae et~al.(2024)Bae, Fisch, Harutyunyan, Ji, Kim, and Schuster]{bae2024relaxed}
S.~Bae, A.~Fisch, H.~Harutyunyan, Z.~Ji, S.~Kim, and T.~Schuster.
\newblock Relaxed recursive transformers: Effective parameter sharing with layer-wise lora.
\newblock \emph{arXiv preprint arXiv:2410.20672}, 2024.

\bibitem[Ben~Allal et~al.(2024)Ben~Allal, Lozhkov, Penedo, Wolf, and von Werra]{benallal2024cosmopedia}
L.~Ben~Allal, A.~Lozhkov, G.~Penedo, T.~Wolf, and L.~von Werra.
\newblock Cosmopedia, February 2024.
\newblock URL \url{https://huggingface.co/datasets/HuggingFaceTB/cosmopedia}.

\bibitem[Biran et~al.(2024)Biran, Gottesman, Yang, Geva, and Globerson]{biran2024hoppinglateexploringlimitations}
E.~Biran, D.~Gottesman, S.~Yang, M.~Geva, and A.~Globerson.
\newblock Hopping too late: Exploring the limitations of large language models on multi-hop queries, 2024.
\newblock URL \url{https://arxiv.org/abs/2406.12775}.

\bibitem[Clark et~al.(2018)Clark, Cowhey, Etzioni, Khot, Sabharwal, Schoenick, and Tafjord]{allenai_arc}
P.~Clark, I.~Cowhey, O.~Etzioni, T.~Khot, A.~Sabharwal, C.~Schoenick, and O.~Tafjord.
\newblock Think you have solved question answering? try arc, the ai2 reasoning challenge.
\newblock \emph{arXiv:1803.05457v1}, 2018.

\bibitem[DeepSeek{-}AI et~al.(2024)DeepSeek{-}AI, Liu, Feng, Xue, Wang, Wu, Lu, Zhao, Deng, Zhang, Ruan, Dai, Guo, Yang, Chen, Ji, Li, Lin, Dai, Luo, Hao, Chen, Li, Zhang, Bao, Xu, Wang, Zhang, Ding, Xin, Gao, Li, Qu, Cai, Liang, Guo, Ni, Li, Wang, Chen, Chen, Yuan, Qiu, Li, Song, Dong, Hu, Gao, Guan, Huang, Yu, Wang, Zhang, Xu, Xia, Zhao, Wang, Zhang, Li, Wang, Zhang, Zhang, Tang, Li, Tian, Huang, Wang, Zhang, Wang, Zhu, Chen, Du, Chen, Jin, Ge, Zhang, Pan, Wang, Xu, Zhang, Chen, Li, Lu, Zhou, Chen, Wu, Ye, Ye, Ma, Wang, Zhou, Yu, Zhou, Pan, Wang, Yun, Pei, Sun, Xiao, and Zeng]{DBLP:journals/corr/abs-2412-19437}
DeepSeek{-}AI, A.~Liu, B.~Feng, B.~Xue, B.~Wang, B.~Wu, C.~Lu, C.~Zhao, C.~Deng, C.~Zhang, C.~Ruan, D.~Dai, D.~Guo, D.~Yang, D.~Chen, D.~Ji, E.~Li, F.~Lin, F.~Dai, F.~Luo, G.~Hao, G.~Chen, G.~Li, H.~Zhang, H.~Bao, H.~Xu, H.~Wang, H.~Zhang, H.~Ding, H.~Xin, H.~Gao, H.~Li, H.~Qu, J.~L. Cai, J.~Liang, J.~Guo, J.~Ni, J.~Li, J.~Wang, J.~Chen, J.~Chen, J.~Yuan, J.~Qiu, J.~Li, J.~Song, K.~Dong, K.~Hu, K.~Gao, K.~Guan, K.~Huang, K.~Yu, L.~Wang, L.~Zhang, L.~Xu, L.~Xia, L.~Zhao, L.~Wang, L.~Zhang, M.~Li, M.~Wang, M.~Zhang, M.~Zhang, M.~Tang, M.~Li, N.~Tian, P.~Huang, P.~Wang, P.~Zhang, Q.~Wang, Q.~Zhu, Q.~Chen, Q.~Du, R.~J. Chen, R.~L. Jin, R.~Ge, R.~Zhang, R.~Pan, R.~Wang, R.~Xu, R.~Zhang, R.~Chen, S.~S. Li, S.~Lu, S.~Zhou, S.~Chen, S.~Wu, S.~Ye, S.~Ye, S.~Ma, S.~Wang, S.~Zhou, S.~Yu, S.~Zhou, S.~Pan, T.~Wang, T.~Yun, T.~Pei, T.~Sun, W.~L. Xiao, and W.~Zeng.
\newblock Deepseek-v3 technical report.
\newblock \emph{CoRR}, abs/2412.19437, 2024.
\newblock \doi{10.48550/ARXIV.2412.19437}.
\newblock URL \url{https://doi.org/10.48550/arXiv.2412.19437}.

\bibitem[Dehghani et~al.(2019)Dehghani, Gouws, Vinyals, Uszkoreit, and Kaiser]{DBLP:conf/iclr/DehghaniGVUK19}
M.~Dehghani, S.~Gouws, O.~Vinyals, J.~Uszkoreit, and L.~Kaiser.
\newblock Universal transformers.
\newblock In \emph{7th International Conference on Learning Representations, {ICLR} 2019, New Orleans, LA, USA, May 6-9, 2019}. OpenReview.net, 2019.
\newblock URL \url{https://openreview.net/forum?id=HyzdRiR9Y7}.

\bibitem[Du et~al.(2022)Du, Huang, Dai, Tong, Lepikhin, Xu, Krikun, Zhou, Yu, Firat, Zoph, Fedus, Bosma, Zhou, Wang, Wang, Webster, Pellat, Robinson, Meier-Hellstern, Duke, Dixon, Zhang, Le, Wu, Chen, and Cui]{du2022glamefficientscalinglanguage}
N.~Du, Y.~Huang, A.~M. Dai, S.~Tong, D.~Lepikhin, Y.~Xu, M.~Krikun, Y.~Zhou, A.~W. Yu, O.~Firat, B.~Zoph, L.~Fedus, M.~Bosma, Z.~Zhou, T.~Wang, Y.~E. Wang, K.~Webster, M.~Pellat, K.~Robinson, K.~Meier-Hellstern, T.~Duke, L.~Dixon, K.~Zhang, Q.~V. Le, Y.~Wu, Z.~Chen, and C.~Cui.
\newblock Glam: Efficient scaling of language models with mixture-of-experts, 2022.
\newblock URL \url{https://arxiv.org/abs/2112.06905}.

\bibitem[Elfwing et~al.(2017)Elfwing, Uchibe, and Doya]{elfwing2017sigmoidweightedlinearunitsneural}
S.~Elfwing, E.~Uchibe, and K.~Doya.
\newblock Sigmoid-weighted linear units for neural network function approximation in reinforcement learning, 2017.
\newblock URL \url{https://arxiv.org/abs/1702.03118}.

\bibitem[Elhage et~al.(2022)Elhage, Hume, Olsson, Nanda, Henighan, Johnston, ElShowk, Joseph, DasSarma, Mann, Hernandez, Askell, Ndousse, Jones, Drain, Chen, Bai, Ganguli, Lovitt, Hatfield-Dodds, Kernion, Conerly, Kravec, Fort, Kadavath, Jacobson, Tran-Johnson, Kaplan, Clark, Brown, McCandlish, Amodei, and Olah]{elhage2022solu}
N.~Elhage, T.~Hume, C.~Olsson, N.~Nanda, T.~Henighan, S.~Johnston, S.~ElShowk, N.~Joseph, N.~DasSarma, B.~Mann, D.~Hernandez, A.~Askell, K.~Ndousse, A.~Jones, D.~Drain, A.~Chen, Y.~Bai, D.~Ganguli, L.~Lovitt, Z.~Hatfield-Dodds, J.~Kernion, T.~Conerly, S.~Kravec, S.~Fort, S.~Kadavath, J.~Jacobson, E.~Tran-Johnson, J.~Kaplan, J.~Clark, T.~Brown, S.~McCandlish, D.~Amodei, and C.~Olah.
\newblock Softmax linear units.
\newblock \emph{Transformer Circuits Thread}, 2022.
\newblock https://transformer-circuits.pub/2022/solu/index.html.

\bibitem[Elhoushi et~al.(2024)Elhoushi, Shrivastava, Liskovich, Hosmer, Wasti, Lai, Mahmoud, Acun, Agarwal, Roman, Aly, Chen, and Wu]{DBLP:conf/acl/ElhoushiSLHWL0A24}
M.~Elhoushi, A.~Shrivastava, D.~Liskovich, B.~Hosmer, B.~Wasti, L.~Lai, A.~Mahmoud, B.~Acun, S.~Agarwal, A.~Roman, A.~A. Aly, B.~Chen, and C.~Wu.
\newblock Layerskip: Enabling early exit inference and self-speculative decoding.
\newblock In L.~Ku, A.~Martins, and V.~Srikumar, editors, \emph{Proceedings of the 62nd Annual Meeting of the Association for Computational Linguistics (Volume 1: Long Papers), {ACL} 2024, Bangkok, Thailand, August 11-16, 2024}, pages 12622--12642. Association for Computational Linguistics, 2024.
\newblock \doi{10.18653/V1/2024.ACL-LONG.681}.
\newblock URL \url{https://doi.org/10.18653/v1/2024.acl-long.681}.

\bibitem[Fan et~al.(2019)Fan, Grave, and Joulin]{fan2019reducing}
A.~Fan, E.~Grave, and A.~Joulin.
\newblock Reducing transformer depth on demand with structured dropout.
\newblock \emph{arXiv preprint arXiv:1909.11556}, 2019.

\bibitem[Fedus et~al.(2022)Fedus, Zoph, and Shazeer]{fedus2022switch}
W.~Fedus, B.~Zoph, and N.~Shazeer.
\newblock Switch transformers: Scaling to trillion parameter models with simple and efficient sparsity.
\newblock \emph{Journal of Machine Learning Research}, 23\penalty0 (120):\penalty0 1--39, 2022.

\bibitem[Geiping et~al.(2025)Geiping, McLeish, Jain, Kirchenbauer, Singh, Bartoldson, Kailkhura, Bhatele, and Goldstein]{geiping2025scalingtesttimecomputelatent}
J.~Geiping, S.~McLeish, N.~Jain, J.~Kirchenbauer, S.~Singh, B.~R. Bartoldson, B.~Kailkhura, A.~Bhatele, and T.~Goldstein.
\newblock Scaling up test-time compute with latent reasoning: A recurrent depth approach, 2025.
\newblock URL \url{https://arxiv.org/abs/2502.05171}.

\bibitem[Giannou et~al.(2023)Giannou, Rajput, Sohn, Lee, Lee, and Papailiopoulos]{DBLP:conf/icml/GiannouRS0LP23}
A.~Giannou, S.~Rajput, J.~Sohn, K.~Lee, J.~D. Lee, and D.~Papailiopoulos.
\newblock Looped transformers as programmable computers.
\newblock In A.~Krause, E.~Brunskill, K.~Cho, B.~Engelhardt, S.~Sabato, and J.~Scarlett, editors, \emph{International Conference on Machine Learning, {ICML} 2023, 23-29 July 2023, Honolulu, Hawaii, {USA}}, volume 202 of \emph{Proceedings of Machine Learning Research}, pages 11398--11442. {PMLR}, 2023.
\newblock URL \url{https://proceedings.mlr.press/v202/giannou23a.html}.

\bibitem[Grattafiori et~al.(2024)Grattafiori, Dubey, Jauhri, Pandey, Kadian, Al-Dahle, Letman, Mathur, Schelten, Vaughan, Yang, Fan, Goyal, Hartshorn, Yang, Mitra, Sravankumar, Korenev, Hinsvark, Rao, Zhang, Rodriguez, Gregerson, Spataru, Roziere, Biron, Tang, Chern, Caucheteux, Nayak, Bi, Marra, McConnell, Keller, Touret, Wu, Wong, Ferrer, Nikolaidis, Allonsius, Song, Pintz, Livshits, Wyatt, Esiobu, Choudhary, Mahajan, Garcia-Olano, Perino, Hupkes, Lakomkin, AlBadawy, Lobanova, Dinan, Smith, Radenovic, Guzmán, Zhang, Synnaeve, Lee, Anderson, Thattai, Nail, Mialon, Pang, Cucurell, Nguyen, Korevaar, Xu, Touvron, Zarov, Ibarra, Kloumann, Misra, Evtimov, Zhang, Copet, Lee, Geffert, Vranes, Park, Mahadeokar, Shah, van~der Linde, Billock, Hong, Lee, Fu, Chi, Huang, Liu, Wang, Yu, Bitton, Spisak, Park, Rocca, Johnstun, Saxe, Jia, Alwala, Prasad, Upasani, Plawiak, Li, Heafield, Stone, El-Arini, Iyer, Malik, Chiu, Bhalla, Lakhotia, Rantala-Yeary, van~der Maaten, Chen, Tan, Jenkins, Martin, Madaan, Malo, Blecher,
  Landzaat, de~Oliveira, Muzzi, Pasupuleti, Singh, Paluri, Kardas, Tsimpoukelli, Oldham, Rita, Pavlova, Kambadur, Lewis, Si, Singh, Hassan, Goyal, Torabi, Bashlykov, Bogoychev, Chatterji, Zhang, Duchenne, Çelebi, Alrassy, Zhang, Li, Vasic, Weng, Bhargava, Dubal, Krishnan, Koura, Xu, He, Dong, Srinivasan, Ganapathy, Calderer, Cabral, Stojnic, Raileanu, Maheswari, Girdhar, Patel, Sauvestre, Polidoro, Sumbaly, Taylor, Silva, Hou, Wang, Hosseini, Chennabasappa, Singh, Bell, Kim, Edunov, Nie, Narang, Raparthy, Shen, Wan, Bhosale, Zhang, Vandenhende, Batra, Whitman, Sootla, Collot, Gururangan, Borodinsky, Herman, Fowler, Sheasha, Georgiou, Scialom, Speckbacher, Mihaylov, Xiao, Karn, Goswami, Gupta, Ramanathan, Kerkez, Gonguet, Do, Vogeti, Albiero, Petrovic, Chu, Xiong, Fu, Meers, Martinet, Wang, Wang, Tan, Xia, Xie, Jia, Wang, Goldschlag, Gaur, Babaei, Wen, Song, Zhang, Li, Mao, Coudert, Yan, Chen, Papakipos, Singh, Srivastava, Jain, Kelsey, Shajnfeld, Gangidi, Victoria, Goldstand, Menon, Sharma, Boesenberg,
  Baevski, Feinstein, Kallet, Sangani, Teo, Yunus, Lupu, Alvarado, Caples, Gu, Ho, Poulton, Ryan, Ramchandani, Dong, Franco, Goyal, Saraf, Chowdhury, Gabriel, Bharambe, Eisenman, Yazdan, James, Maurer, Leonhardi, Huang, Loyd, Paola, Paranjape, Liu, Wu, Ni, Hancock, Wasti, Spence, Stojkovic, Gamido, Montalvo, Parker, Burton, Mejia, Liu, Wang, Kim, Zhou, Hu, Chu, Cai, Tindal, Feichtenhofer, Gao, Civin, Beaty, Kreymer, Li, Adkins, Xu, Testuggine, David, Parikh, Liskovich, Foss, Wang, Le, Holland, Dowling, Jamil, Montgomery, Presani, Hahn, Wood, Le, Brinkman, Arcaute, Dunbar, Smothers, Sun, Kreuk, Tian, Kokkinos, Ozgenel, Caggioni, Kanayet, Seide, Florez, Schwarz, Badeer, Swee, Halpern, Herman, Sizov, Guangyi, Zhang, Lakshminarayanan, Inan, Shojanazeri, Zou, Wang, Zha, Habeeb, Rudolph, Suk, Aspegren, Goldman, Zhan, Damlaj, Molybog, Tufanov, Leontiadis, Veliche, Gat, Weissman, Geboski, Kohli, Lam, Asher, Gaya, Marcus, Tang, Chan, Zhen, Reizenstein, Teboul, Zhong, Jin, Yang, Cummings, Carvill, Shepard, McPhie,
  Torres, Ginsburg, Wang, Wu, U, Saxena, Khandelwal, Zand, Matosich, Veeraraghavan, Michelena, Li, Jagadeesh, Huang, Chawla, Huang, Chen, Garg, A, Silva, Bell, Zhang, Guo, Yu, Moshkovich, Wehrstedt, Khabsa, Avalani, Bhatt, Mankus, Hasson, Lennie, Reso, Groshev, Naumov, Lathi, Keneally, Liu, Seltzer, Valko, Restrepo, Patel, Vyatskov, Samvelyan, Clark, Macey, Wang, Hermoso, Metanat, Rastegari, Bansal, Santhanam, Parks, White, Bawa, Singhal, Egebo, Usunier, Mehta, Laptev, Dong, Cheng, Chernoguz, Hart, Salpekar, Kalinli, Kent, Parekh, Saab, Balaji, Rittner, Bontrager, Roux, Dollar, Zvyagina, Ratanchandani, Yuvraj, Liang, Alao, Rodriguez, Ayub, Murthy, Nayani, Mitra, Parthasarathy, Li, Hogan, Battey, Wang, Howes, Rinott, Mehta, Siby, Bondu, Datta, Chugh, Hunt, Dhillon, Sidorov, Pan, Mahajan, Verma, Yamamoto, Ramaswamy, Lindsay, Lindsay, Feng, Lin, Zha, Patil, Shankar, Zhang, Zhang, Wang, Agarwal, Sajuyigbe, Chintala, Max, Chen, Kehoe, Satterfield, Govindaprasad, Gupta, Deng, Cho, Virk, Subramanian, Choudhury,
  Goldman, Remez, Glaser, Best, Koehler, Robinson, Li, Zhang, Matthews, Chou, Shaked, Vontimitta, Ajayi, Montanez, Mohan, Kumar, Mangla, Ionescu, Poenaru, Mihailescu, Ivanov, Li, Wang, Jiang, Bouaziz, Constable, Tang, Wu, Wang, Wu, Gao, Kleinman, Chen, Hu, Jia, Qi, Li, Zhang, Zhang, Adi, Nam, Yu, Wang, Zhao, Hao, Qian, Li, He, Rait, DeVito, Rosnbrick, Wen, Yang, Zhao, and Ma]{grattafiori2024llama3herdmodels}
A.~Grattafiori, A.~Dubey, A.~Jauhri, A.~Pandey, A.~Kadian, A.~Al-Dahle, A.~Letman, A.~Mathur, A.~Schelten, A.~Vaughan, A.~Yang, A.~Fan, A.~Goyal, A.~Hartshorn, A.~Yang, A.~Mitra, A.~Sravankumar, A.~Korenev, A.~Hinsvark, A.~Rao, A.~Zhang, A.~Rodriguez, A.~Gregerson, A.~Spataru, B.~Roziere, B.~Biron, B.~Tang, B.~Chern, C.~Caucheteux, C.~Nayak, C.~Bi, C.~Marra, C.~McConnell, C.~Keller, C.~Touret, C.~Wu, C.~Wong, C.~C. Ferrer, C.~Nikolaidis, D.~Allonsius, D.~Song, D.~Pintz, D.~Livshits, D.~Wyatt, D.~Esiobu, D.~Choudhary, D.~Mahajan, D.~Garcia-Olano, D.~Perino, D.~Hupkes, E.~Lakomkin, E.~AlBadawy, E.~Lobanova, E.~Dinan, E.~M. Smith, F.~Radenovic, F.~Guzmán, F.~Zhang, G.~Synnaeve, G.~Lee, G.~L. Anderson, G.~Thattai, G.~Nail, G.~Mialon, G.~Pang, G.~Cucurell, H.~Nguyen, H.~Korevaar, H.~Xu, H.~Touvron, I.~Zarov, I.~A. Ibarra, I.~Kloumann, I.~Misra, I.~Evtimov, J.~Zhang, J.~Copet, J.~Lee, J.~Geffert, J.~Vranes, J.~Park, J.~Mahadeokar, J.~Shah, J.~van~der Linde, J.~Billock, J.~Hong, J.~Lee, J.~Fu, J.~Chi, J.~Huang,
  J.~Liu, J.~Wang, J.~Yu, J.~Bitton, J.~Spisak, J.~Park, J.~Rocca, J.~Johnstun, J.~Saxe, J.~Jia, K.~V. Alwala, K.~Prasad, K.~Upasani, K.~Plawiak, K.~Li, K.~Heafield, K.~Stone, K.~El-Arini, K.~Iyer, K.~Malik, K.~Chiu, K.~Bhalla, K.~Lakhotia, L.~Rantala-Yeary, L.~van~der Maaten, L.~Chen, L.~Tan, L.~Jenkins, L.~Martin, L.~Madaan, L.~Malo, L.~Blecher, L.~Landzaat, L.~de~Oliveira, M.~Muzzi, M.~Pasupuleti, M.~Singh, M.~Paluri, M.~Kardas, M.~Tsimpoukelli, M.~Oldham, M.~Rita, M.~Pavlova, M.~Kambadur, M.~Lewis, M.~Si, M.~K. Singh, M.~Hassan, N.~Goyal, N.~Torabi, N.~Bashlykov, N.~Bogoychev, N.~Chatterji, N.~Zhang, O.~Duchenne, O.~Çelebi, P.~Alrassy, P.~Zhang, P.~Li, P.~Vasic, P.~Weng, P.~Bhargava, P.~Dubal, P.~Krishnan, P.~S. Koura, P.~Xu, Q.~He, Q.~Dong, R.~Srinivasan, R.~Ganapathy, R.~Calderer, R.~S. Cabral, R.~Stojnic, R.~Raileanu, R.~Maheswari, R.~Girdhar, R.~Patel, R.~Sauvestre, R.~Polidoro, R.~Sumbaly, R.~Taylor, R.~Silva, R.~Hou, R.~Wang, S.~Hosseini, S.~Chennabasappa, S.~Singh, S.~Bell, S.~S. Kim, S.~Edunov,
  S.~Nie, S.~Narang, S.~Raparthy, S.~Shen, S.~Wan, S.~Bhosale, S.~Zhang, S.~Vandenhende, S.~Batra, S.~Whitman, S.~Sootla, S.~Collot, S.~Gururangan, S.~Borodinsky, T.~Herman, T.~Fowler, T.~Sheasha, T.~Georgiou, T.~Scialom, T.~Speckbacher, T.~Mihaylov, T.~Xiao, U.~Karn, V.~Goswami, V.~Gupta, V.~Ramanathan, V.~Kerkez, V.~Gonguet, V.~Do, V.~Vogeti, V.~Albiero, V.~Petrovic, W.~Chu, W.~Xiong, W.~Fu, W.~Meers, X.~Martinet, X.~Wang, X.~Wang, X.~E. Tan, X.~Xia, X.~Xie, X.~Jia, X.~Wang, Y.~Goldschlag, Y.~Gaur, Y.~Babaei, Y.~Wen, Y.~Song, Y.~Zhang, Y.~Li, Y.~Mao, Z.~D. Coudert, Z.~Yan, Z.~Chen, Z.~Papakipos, A.~Singh, A.~Srivastava, A.~Jain, A.~Kelsey, A.~Shajnfeld, A.~Gangidi, A.~Victoria, A.~Goldstand, A.~Menon, A.~Sharma, A.~Boesenberg, A.~Baevski, A.~Feinstein, A.~Kallet, A.~Sangani, A.~Teo, A.~Yunus, A.~Lupu, A.~Alvarado, A.~Caples, A.~Gu, A.~Ho, A.~Poulton, A.~Ryan, A.~Ramchandani, A.~Dong, A.~Franco, A.~Goyal, A.~Saraf, A.~Chowdhury, A.~Gabriel, A.~Bharambe, A.~Eisenman, A.~Yazdan, B.~James, B.~Maurer,
  B.~Leonhardi, B.~Huang, B.~Loyd, B.~D. Paola, B.~Paranjape, B.~Liu, B.~Wu, B.~Ni, B.~Hancock, B.~Wasti, B.~Spence, B.~Stojkovic, B.~Gamido, B.~Montalvo, C.~Parker, C.~Burton, C.~Mejia, C.~Liu, C.~Wang, C.~Kim, C.~Zhou, C.~Hu, C.-H. Chu, C.~Cai, C.~Tindal, C.~Feichtenhofer, C.~Gao, D.~Civin, D.~Beaty, D.~Kreymer, D.~Li, D.~Adkins, D.~Xu, D.~Testuggine, D.~David, D.~Parikh, D.~Liskovich, D.~Foss, D.~Wang, D.~Le, D.~Holland, E.~Dowling, E.~Jamil, E.~Montgomery, E.~Presani, E.~Hahn, E.~Wood, E.-T. Le, E.~Brinkman, E.~Arcaute, E.~Dunbar, E.~Smothers, F.~Sun, F.~Kreuk, F.~Tian, F.~Kokkinos, F.~Ozgenel, F.~Caggioni, F.~Kanayet, F.~Seide, G.~M. Florez, G.~Schwarz, G.~Badeer, G.~Swee, G.~Halpern, G.~Herman, G.~Sizov, Guangyi, Zhang, G.~Lakshminarayanan, H.~Inan, H.~Shojanazeri, H.~Zou, H.~Wang, H.~Zha, H.~Habeeb, H.~Rudolph, H.~Suk, H.~Aspegren, H.~Goldman, H.~Zhan, I.~Damlaj, I.~Molybog, I.~Tufanov, I.~Leontiadis, I.-E. Veliche, I.~Gat, J.~Weissman, J.~Geboski, J.~Kohli, J.~Lam, J.~Asher, J.-B. Gaya, J.~Marcus,
  J.~Tang, J.~Chan, J.~Zhen, J.~Reizenstein, J.~Teboul, J.~Zhong, J.~Jin, J.~Yang, J.~Cummings, J.~Carvill, J.~Shepard, J.~McPhie, J.~Torres, J.~Ginsburg, J.~Wang, K.~Wu, K.~H. U, K.~Saxena, K.~Khandelwal, K.~Zand, K.~Matosich, K.~Veeraraghavan, K.~Michelena, K.~Li, K.~Jagadeesh, K.~Huang, K.~Chawla, K.~Huang, L.~Chen, L.~Garg, L.~A, L.~Silva, L.~Bell, L.~Zhang, L.~Guo, L.~Yu, L.~Moshkovich, L.~Wehrstedt, M.~Khabsa, M.~Avalani, M.~Bhatt, M.~Mankus, M.~Hasson, M.~Lennie, M.~Reso, M.~Groshev, M.~Naumov, M.~Lathi, M.~Keneally, M.~Liu, M.~L. Seltzer, M.~Valko, M.~Restrepo, M.~Patel, M.~Vyatskov, M.~Samvelyan, M.~Clark, M.~Macey, M.~Wang, M.~J. Hermoso, M.~Metanat, M.~Rastegari, M.~Bansal, N.~Santhanam, N.~Parks, N.~White, N.~Bawa, N.~Singhal, N.~Egebo, N.~Usunier, N.~Mehta, N.~P. Laptev, N.~Dong, N.~Cheng, O.~Chernoguz, O.~Hart, O.~Salpekar, O.~Kalinli, P.~Kent, P.~Parekh, P.~Saab, P.~Balaji, P.~Rittner, P.~Bontrager, P.~Roux, P.~Dollar, P.~Zvyagina, P.~Ratanchandani, P.~Yuvraj, Q.~Liang, R.~Alao, R.~Rodriguez,
  R.~Ayub, R.~Murthy, R.~Nayani, R.~Mitra, R.~Parthasarathy, R.~Li, R.~Hogan, R.~Battey, R.~Wang, R.~Howes, R.~Rinott, S.~Mehta, S.~Siby, S.~J. Bondu, S.~Datta, S.~Chugh, S.~Hunt, S.~Dhillon, S.~Sidorov, S.~Pan, S.~Mahajan, S.~Verma, S.~Yamamoto, S.~Ramaswamy, S.~Lindsay, S.~Lindsay, S.~Feng, S.~Lin, S.~C. Zha, S.~Patil, S.~Shankar, S.~Zhang, S.~Zhang, S.~Wang, S.~Agarwal, S.~Sajuyigbe, S.~Chintala, S.~Max, S.~Chen, S.~Kehoe, S.~Satterfield, S.~Govindaprasad, S.~Gupta, S.~Deng, S.~Cho, S.~Virk, S.~Subramanian, S.~Choudhury, S.~Goldman, T.~Remez, T.~Glaser, T.~Best, T.~Koehler, T.~Robinson, T.~Li, T.~Zhang, T.~Matthews, T.~Chou, T.~Shaked, V.~Vontimitta, V.~Ajayi, V.~Montanez, V.~Mohan, V.~S. Kumar, V.~Mangla, V.~Ionescu, V.~Poenaru, V.~T. Mihailescu, V.~Ivanov, W.~Li, W.~Wang, W.~Jiang, W.~Bouaziz, W.~Constable, X.~Tang, X.~Wu, X.~Wang, X.~Wu, X.~Gao, Y.~Kleinman, Y.~Chen, Y.~Hu, Y.~Jia, Y.~Qi, Y.~Li, Y.~Zhang, Y.~Zhang, Y.~Adi, Y.~Nam, Yu, Wang, Y.~Zhao, Y.~Hao, Y.~Qian, Y.~Li, Y.~He, Z.~Rait, Z.~DeVito,
  Z.~Rosnbrick, Z.~Wen, Z.~Yang, Z.~Zhao, and Z.~Ma.
\newblock The llama 3 herd of models, 2024.
\newblock URL \url{https://arxiv.org/abs/2407.21783}.

\bibitem[Gu and Dao(2023)]{gu2023mamba}
A.~Gu and T.~Dao.
\newblock Mamba: Linear-time sequence modeling with selective state spaces.
\newblock \emph{arXiv preprint arXiv:2312.00752}, 2023.

\bibitem[Gurnee et~al.(2023)Gurnee, Nanda, Pauly, Harvey, Troitskii, and Bertsimas]{DBLP:journals/tmlr/GurneeNPHTB23}
W.~Gurnee, N.~Nanda, M.~Pauly, K.~Harvey, D.~Troitskii, and D.~Bertsimas.
\newblock Finding neurons in a haystack: Case studies with sparse probing.
\newblock \emph{Trans. Mach. Learn. Res.}, 2023, 2023.
\newblock URL \url{https://openreview.net/forum?id=JYs1R9IMJr}.

\bibitem[Hao et~al.(2024)Hao, Sukhbaatar, Su, Li, Hu, Weston, and Tian]{DBLP:journals/corr/abs-2412-06769}
S.~Hao, S.~Sukhbaatar, D.~Su, X.~Li, Z.~Hu, J.~Weston, and Y.~Tian.
\newblock Training large language models to reason in a continuous latent space.
\newblock \emph{CoRR}, abs/2412.06769, 2024.
\newblock \doi{10.48550/ARXIV.2412.06769}.
\newblock URL \url{https://doi.org/10.48550/arXiv.2412.06769}.

\bibitem[Hoffmann et~al.(2022)Hoffmann, Borgeaud, Mensch, Buchatskaya, Cai, Rutherford, de~Las~Casas, Hendricks, Welbl, Clark, Hennigan, Noland, Millican, van~den Driessche, Damoc, Guy, Osindero, Simonyan, Elsen, Rae, Vinyals, and Sifre]{DBLP:journals/corr/abs-2203-15556}
J.~Hoffmann, S.~Borgeaud, A.~Mensch, E.~Buchatskaya, T.~Cai, E.~Rutherford, D.~de~Las~Casas, L.~A. Hendricks, J.~Welbl, A.~Clark, T.~Hennigan, E.~Noland, K.~Millican, G.~van~den Driessche, B.~Damoc, A.~Guy, S.~Osindero, K.~Simonyan, E.~Elsen, J.~W. Rae, O.~Vinyals, and L.~Sifre.
\newblock Training compute-optimal large language models.
\newblock \emph{CoRR}, abs/2203.15556, 2022.
\newblock \doi{10.48550/ARXIV.2203.15556}.
\newblock URL \url{https://doi.org/10.48550/arXiv.2203.15556}.

\bibitem[Hsieh et~al.(2023)Hsieh, Li, Yeh, Nakhost, Fujii, Ratner, Krishna, Lee, and Pfister]{hsieh2023distilling}
C.-Y. Hsieh, C.-L. Li, C.-K. Yeh, H.~Nakhost, Y.~Fujii, A.~Ratner, R.~Krishna, C.-Y. Lee, and T.~Pfister.
\newblock Distilling step-by-step! outperforming larger language models with less training data and smaller model sizes.
\newblock \emph{arXiv preprint arXiv:2305.02301}, 2023.

\bibitem[Hägele et~al.(2024)Hägele, Bakouch, Kosson, Allal, Werra, and Jaggi]{hägele2024scalinglawscomputeoptimaltraining}
A.~Hägele, E.~Bakouch, A.~Kosson, L.~B. Allal, L.~V. Werra, and M.~Jaggi.
\newblock Scaling laws and compute-optimal training beyond fixed training durations, 2024.
\newblock URL \url{https://arxiv.org/abs/2405.18392}.

\bibitem[Jumper et~al.(2021)Jumper, Evans, Pritzel, Green, Figurnov, Ronneberger, Tunyasuvunakool, Bates, Žídek, Potapenko, Bridgland, Meyer, Kohl, Ballard, Cowie, Romera-Paredes, Nikolov, Jain, Adler, Back, Petersen, Reiman, Clancy, Zielinski, Steinegger, Pacholska, Berghammer, Bodenstein, Silver, Vinyals, Senior, Kavukcuoglu, Kohli, and Hassabis]{jumper_highly_2021}
J.~Jumper, R.~Evans, A.~Pritzel, T.~Green, M.~Figurnov, O.~Ronneberger, K.~Tunyasuvunakool, R.~Bates, A.~Žídek, A.~Potapenko, A.~Bridgland, C.~Meyer, S.~A.~A. Kohl, A.~J. Ballard, A.~Cowie, B.~Romera-Paredes, S.~Nikolov, R.~Jain, J.~Adler, T.~Back, S.~Petersen, D.~Reiman, E.~Clancy, M.~Zielinski, M.~Steinegger, M.~Pacholska, T.~Berghammer, S.~Bodenstein, D.~Silver, O.~Vinyals, A.~W. Senior, K.~Kavukcuoglu, P.~Kohli, and D.~Hassabis.
\newblock Highly accurate protein structure prediction with {AlphaFold}.
\newblock \emph{Nature}, 596\penalty0 (7873):\penalty0 583--589, Aug. 2021.
\newblock ISSN 1476-4687.
\newblock \doi{10.1038/s41586-021-03819-2}.
\newblock URL \url{https://doi.org/10.1038/s41586-021-03819-2}.

\bibitem[Kaplan et~al.(2024)Kaplan, Oren, Reif, and Schwartz]{DBLP:journals/corr/abs-2410-05864}
G.~Kaplan, M.~Oren, Y.~Reif, and R.~Schwartz.
\newblock From tokens to words: On the inner lexicon of llms.
\newblock \emph{CoRR}, abs/2410.05864, 2024.
\newblock \doi{10.48550/ARXIV.2410.05864}.
\newblock URL \url{https://doi.org/10.48550/arXiv.2410.05864}.

\bibitem[Kaplan et~al.(2020)Kaplan, McCandlish, Henighan, Brown, Chess, Child, Gray, Radford, Wu, and Amodei]{DBLP:journals/corr/abs-2001-08361}
J.~Kaplan, S.~McCandlish, T.~Henighan, T.~B. Brown, B.~Chess, R.~Child, S.~Gray, A.~Radford, J.~Wu, and D.~Amodei.
\newblock Scaling laws for neural language models.
\newblock \emph{CoRR}, abs/2001.08361, 2020.
\newblock URL \url{https://arxiv.org/abs/2001.08361}.

\bibitem[Katz and Belinkov(2023)]{DBLP:conf/emnlp/KatzB23}
S.~Katz and Y.~Belinkov.
\newblock {VISIT:} visualizing and interpreting the semantic information flow of transformers.
\newblock In H.~Bouamor, J.~Pino, and K.~Bali, editors, \emph{Findings of the Association for Computational Linguistics: {EMNLP} 2023, Singapore, December 6-10, 2023}, pages 14094--14113. Association for Computational Linguistics, 2023.
\newblock \doi{10.18653/V1/2023.FINDINGS-EMNLP.939}.
\newblock URL \url{https://doi.org/10.18653/v1/2023.findings-emnlp.939}.

\bibitem[Kocetkov et~al.(2022)Kocetkov, Li, Ben~Allal, Li, Mou, Muñoz~Ferrandis, Jernite, Mitchell, Hughes, Wolf, Bahdanau, von Werra, and de~Vries]{Kocetkov2022TheStack}
D.~Kocetkov, R.~Li, L.~Ben~Allal, J.~Li, C.~Mou, C.~Muñoz~Ferrandis, Y.~Jernite, M.~Mitchell, S.~Hughes, T.~Wolf, D.~Bahdanau, L.~von Werra, and H.~de~Vries.
\newblock The stack: 3 tb of permissively licensed source code.
\newblock \emph{Preprint}, 2022.

\bibitem[Lu et~al.(2024)Lu, Zhou, Liu, Wang, Mahoney, and Yang]{lu2024alphapruning}
H.~Lu, Y.~Zhou, S.~Liu, Z.~Wang, M.~W. Mahoney, and Y.~Yang.
\newblock Alphapruning: Using heavy-tailed self regularization theory for improved layer-wise pruning of large language models.
\newblock \emph{Advances in Neural Information Processing Systems}, 37:\penalty0 9117--9152, 2024.

\bibitem[McCandlish et~al.(2018)McCandlish, Kaplan, Amodei, and Team]{mccandlish2018empiricalmodellargebatchtraining}
S.~McCandlish, J.~Kaplan, D.~Amodei, and O.~D. Team.
\newblock An empirical model of large-batch training, 2018.
\newblock URL \url{https://arxiv.org/abs/1812.06162}.

\bibitem[Nye et~al.(2021)Nye, Andreassen, Gur{-}Ari, Michalewski, Austin, Bieber, Dohan, Lewkowycz, Bosma, Luan, Sutton, and Odena]{DBLP:journals/corr/abs-2112-00114}
M.~I. Nye, A.~J. Andreassen, G.~Gur{-}Ari, H.~Michalewski, J.~Austin, D.~Bieber, D.~Dohan, A.~Lewkowycz, M.~Bosma, D.~Luan, C.~Sutton, and A.~Odena.
\newblock Show your work: Scratchpads for intermediate computation with language models.
\newblock \emph{CoRR}, abs/2112.00114, 2021.
\newblock URL \url{https://arxiv.org/abs/2112.00114}.

\bibitem[Paperno et~al.(2016)Paperno, Kruszewski, Lazaridou, Pham, Bernardi, Pezzelle, Baroni, Boleda, and Fernández]{paperno2016lambadadatasetwordprediction}
D.~Paperno, G.~Kruszewski, A.~Lazaridou, Q.~N. Pham, R.~Bernardi, S.~Pezzelle, M.~Baroni, G.~Boleda, and R.~Fernández.
\newblock The lambada dataset: Word prediction requiring a broad discourse context, 2016.
\newblock URL \url{https://arxiv.org/abs/1606.06031}.

\bibitem[Penedo et~al.(2024{\natexlab{a}})Penedo, Kydl{\'{\i}}cek, Allal, Lozhkov, Mitchell, Raffel, von Werra, and Wolf]{DBLP:conf/nips/PenedoKALMRW024}
G.~Penedo, H.~Kydl{\'{\i}}cek, L.~B. Allal, A.~Lozhkov, M.~Mitchell, C.~A. Raffel, L.~von Werra, and T.~Wolf.
\newblock The fineweb datasets: Decanting the web for the finest text data at scale.
\newblock In A.~Globersons, L.~Mackey, D.~Belgrave, A.~Fan, U.~Paquet, J.~M. Tomczak, and C.~Zhang, editors, \emph{Advances in Neural Information Processing Systems 38: Annual Conference on Neural Information Processing Systems 2024, NeurIPS 2024, Vancouver, BC, Canada, December 10 - 15, 2024}, 2024{\natexlab{a}}.
\newblock URL \url{http://papers.nips.cc/paper\_files/paper/2024/hash/370df50ccfdf8bde18f8f9c2d9151bda-Abstract-Datasets\_and\_Benchmarks\_Track.html}.

\bibitem[Penedo et~al.(2024{\natexlab{b}})Penedo, Kydlíček, allal, Lozhkov, Mitchell, Raffel, Werra, and Wolf]{penedo2024finewebdatasetsdecantingweb}
G.~Penedo, H.~Kydlíček, L.~B. allal, A.~Lozhkov, M.~Mitchell, C.~Raffel, L.~V. Werra, and T.~Wolf.
\newblock The fineweb datasets: Decanting the web for the finest text data at scale, 2024{\natexlab{b}}.
\newblock URL \url{https://arxiv.org/abs/2406.17557}.

\bibitem[Peng et~al.()Peng, Alcaide, Anthony, Albalak, Arcadinho, Biderman, Cao, Cheng, Chung, Derczynski, et~al.]{peng2023rwkv}
B.~Peng, E.~Alcaide, Q.~G. Anthony, A.~Albalak, S.~Arcadinho, S.~Biderman, H.~Cao, X.~Cheng, M.~N. Chung, L.~Derczynski, et~al.
\newblock Rwkv: Reinventing rnns for the transformer era.
\newblock In \emph{The 2023 Conference on Empirical Methods in Natural Language Processing}.

\bibitem[Pope et~al.(2022)Pope, Douglas, Chowdhery, Devlin, Bradbury, Levskaya, Heek, Xiao, Agrawal, and Dean]{pope2022efficientlyscalingtransformerinference}
R.~Pope, S.~Douglas, A.~Chowdhery, J.~Devlin, J.~Bradbury, A.~Levskaya, J.~Heek, K.~Xiao, S.~Agrawal, and J.~Dean.
\newblock Efficiently scaling transformer inference, 2022.
\newblock URL \url{https://arxiv.org/abs/2211.05102}.

\bibitem[Radford et~al.(2019)Radford, Wu, Child, Luan, Amodei, and Sutskever]{radford2019language}
A.~Radford, J.~Wu, R.~Child, D.~Luan, D.~Amodei, and I.~Sutskever.
\newblock Language models are unsupervised multitask learners.
\newblock \emph{OpenAI}, 2019.
\newblock URL \url{https://cdn.openai.com/better-language-models/language_models_are_unsupervised_multitask_learners.pdf}.
\newblock Accessed: 2024-11-15.

\bibitem[Sardana et~al.(2025)Sardana, Portes, Doubov, and Frankle]{sardana2025chinchillaoptimalaccountinginferencelanguage}
N.~Sardana, J.~Portes, S.~Doubov, and J.~Frankle.
\newblock Beyond chinchilla-optimal: Accounting for inference in language model scaling laws, 2025.
\newblock URL \url{https://arxiv.org/abs/2401.00448}.

\bibitem[Shen et~al.(2025)Shen, Qiu, Kurmanji, Iacob, Sani, Chen, Cancedda, and Lane]{shen2025lunar}
W.~F. Shen, X.~Qiu, M.~Kurmanji, A.~Iacob, L.~Sani, Y.~Chen, N.~Cancedda, and N.~D. Lane.
\newblock Lunar: Llm unlearning via neural activation redirection.
\newblock \emph{arXiv preprint arXiv:2502.07218}, 2025.

\bibitem[Su et~al.(2024)Su, Ahmed, Lu, Pan, Bo, and Liu]{10.1016/j.neucom.2023.127063}
J.~Su, M.~Ahmed, Y.~Lu, S.~Pan, W.~Bo, and Y.~Liu.
\newblock Roformer: Enhanced transformer with rotary position embedding.
\newblock \emph{Neurocomput.}, 568\penalty0 (C), Feb. 2024.
\newblock ISSN 0925-2312.
\newblock \doi{10.1016/j.neucom.2023.127063}.
\newblock URL \url{https://doi.org/10.1016/j.neucom.2023.127063}.

\bibitem[Talmor et~al.(2019)Talmor, Herzig, Lourie, and Berant]{talmor2019commonsenseqaquestionansweringchallenge}
A.~Talmor, J.~Herzig, N.~Lourie, and J.~Berant.
\newblock Commonsenseqa: A question answering challenge targeting commonsense knowledge, 2019.
\newblock URL \url{https://arxiv.org/abs/1811.00937}.

\bibitem[Vaswani et~al.(2017)Vaswani, Shazeer, Parmar, Uszkoreit, Jones, Gomez, Kaiser, and Polosukhin]{DBLP:conf/nips/VaswaniSPUJGKP17}
A.~Vaswani, N.~Shazeer, N.~Parmar, J.~Uszkoreit, L.~Jones, A.~N. Gomez, L.~Kaiser, and I.~Polosukhin.
\newblock Attention is all you need.
\newblock In I.~Guyon, U.~von Luxburg, S.~Bengio, H.~M. Wallach, R.~Fergus, S.~V.~N. Vishwanathan, and R.~Garnett, editors, \emph{Advances in Neural Information Processing Systems 30: Annual Conference on Neural Information Processing Systems 2017, December 4-9, 2017, Long Beach, CA, {USA}}, pages 5998--6008, 2017.
\newblock URL \url{https://proceedings.neurips.cc/paper/2017/hash/3f5ee243547dee91fbd053c1c4a845aa-Abstract.html}.

\bibitem[von Platen and Wolf(2023)]{vonplaten2023smollm}
P.~von Platen and T.~Wolf.
\newblock Smollm: Efficient language models for everyone.
\newblock \url{https://huggingface.co/blog/smollm}, dec 2023.
\newblock Accessed: 2025-05-13.

\bibitem[Wang et~al.(2024)Wang, Zhou, Li, and Huang]{wang20244}
S.~Wang, P.~Zhou, J.~Li, and H.~Huang.
\newblock 4-bit shampoo for memory-efficient network training.
\newblock \emph{Advances in Neural Information Processing Systems}, 37:\penalty0 126997--127029, 2024.

\bibitem[Wolters et~al.(2024)Wolters, Yang, Schlichtmann, and Suzumura]{wolters2024memoryneedoverviewcomputeinmemory}
C.~Wolters, X.~Yang, U.~Schlichtmann, and T.~Suzumura.
\newblock Memory is all you need: An overview of compute-in-memory architectures for accelerating large language model inference, 2024.
\newblock URL \url{https://arxiv.org/abs/2406.08413}.

\bibitem[Xiong et~al.(2020)Xiong, Yang, He, Zheng, Zheng, Xing, Zhang, Lan, Wang, and Liu]{DBLP:conf/icml/XiongYHZZXZLWL20}
R.~Xiong, Y.~Yang, D.~He, K.~Zheng, S.~Zheng, C.~Xing, H.~Zhang, Y.~Lan, L.~Wang, and T.~Liu.
\newblock On layer normalization in the transformer architecture.
\newblock In \emph{Proceedings of the 37th International Conference on Machine Learning, {ICML} 2020, 13-18 July 2020, Virtual Event}, volume 119 of \emph{Proceedings of Machine Learning Research}, pages 10524--10533. {PMLR}, 2020.
\newblock URL \url{http://proceedings.mlr.press/v119/xiong20b.html}.

\bibitem[Yang et~al.(2024)Yang, Lee, Nowak, and Papailiopoulos]{DBLP:conf/iclr/Yang0NP24}
L.~Yang, K.~Lee, R.~D. Nowak, and D.~Papailiopoulos.
\newblock Looped transformers are better at learning learning algorithms.
\newblock In \emph{The Twelfth International Conference on Learning Representations, {ICLR} 2024, Vienna, Austria, May 7-11, 2024}. OpenReview.net, 2024.
\newblock URL \url{https://openreview.net/forum?id=HHbRxoDTxE}.

\bibitem[Zellers et~al.(2019)Zellers, Holtzman, Bisk, Farhadi, and Choi]{zellers2019hellaswagmachinereallyfinish}
R.~Zellers, A.~Holtzman, Y.~Bisk, A.~Farhadi, and Y.~Choi.
\newblock Hellaswag: Can a machine really finish your sentence?, 2019.
\newblock URL \url{https://arxiv.org/abs/1905.07830}.

\bibitem[Zhang and Sennrich(2019)]{zhang2019rootmeansquarelayer}
B.~Zhang and R.~Sennrich.
\newblock Root mean square layer normalization, 2019.
\newblock URL \url{https://arxiv.org/abs/1910.07467}.

\end{thebibliography}

\end{document}